\documentclass[final]{cvpr}

\usepackage{times}
\usepackage{epsfig}
\usepackage{graphicx}
\usepackage{amsmath}
\usepackage{amssymb}
\usepackage{mathtools}

\usepackage{algorithm}
\usepackage{algorithmicx}
\usepackage{algpseudocode}

\usepackage{widetext}

\renewcommand{\thefootnote}{\fnsymbol{footnote}}

\usepackage{xcolor}

\newcommand{\lipsum}[3]{} 

\usepackage[pagebackref=true,breaklinks=true,colorlinks,bookmarks=false]{hyperref}

\def\cvprPaperID{1519} 
\def\confYear{CVPR 2021}

\def\ourmethod{SGF } 

\DeclareMathOperator*{\argmin}{arg\,min}

\begin{document}

\title{Surrogate Gradient Field for Latent Space Manipulation}

\makeatletter
\newcommand{\printfnsymbol}[1]{
  \textsuperscript{\@fnsymbol{#1}}
}
\makeatother

\author{Minjun Li\footnote[1] \quad \quad Yanghua Jin\footnote[1] \quad \quad Huachun Zhu\\
Preferred Networks\\
\tt\small \{minjunli, jinyh, zhu\}@preferred.jp 
}

\maketitle
\footnotetext[1]{Equal contribution.}

\setlength{\wttlinewidth}{0pt}
\begin{widetext}
\vspace{-35pt}
\begin{figure}
    \centering
    \includegraphics[width=0.9\textwidth]{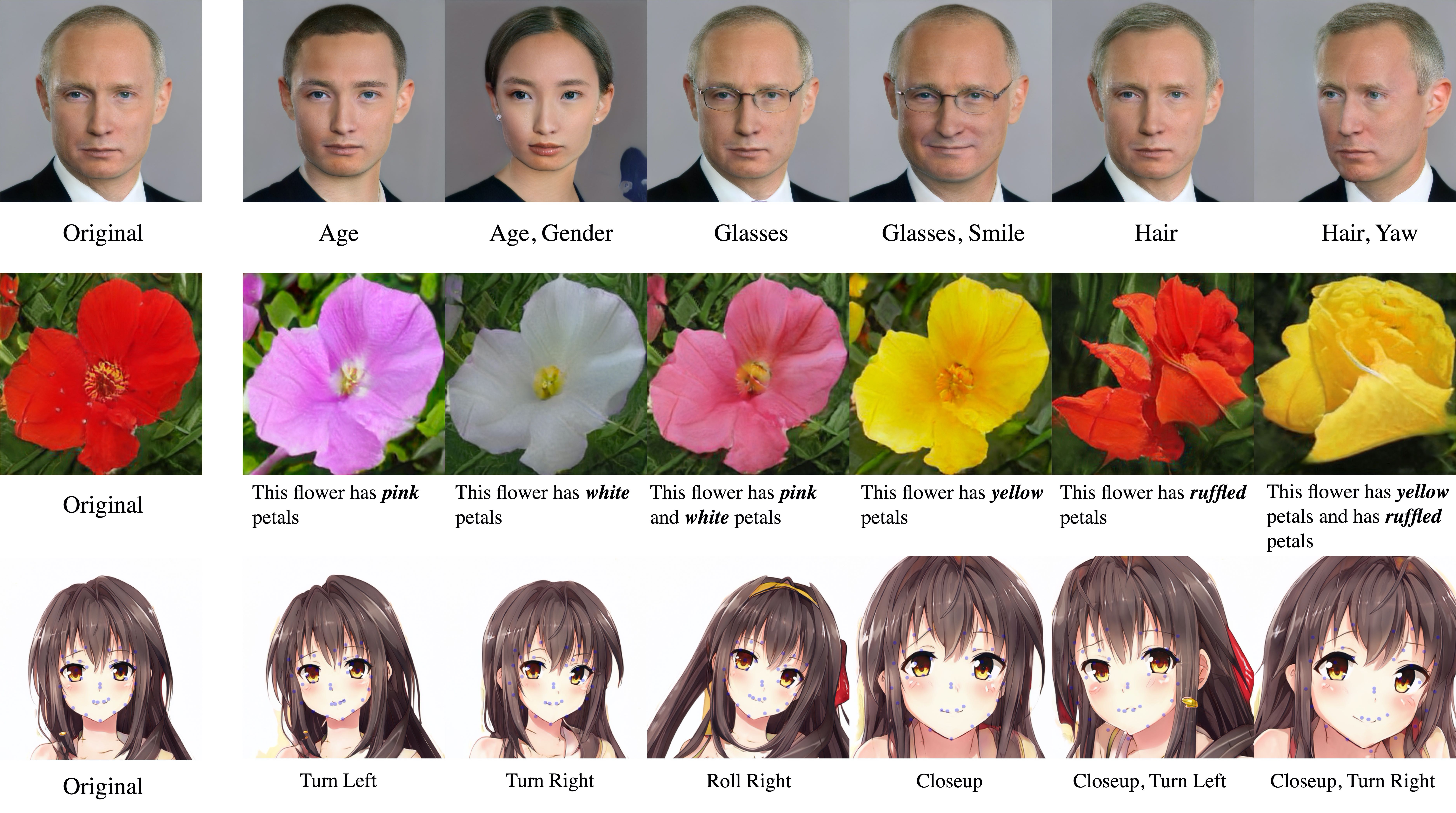}
    \caption{\textbf{
      Our Surrogate Gradient Field (SGF) can edit images with diverse modalities of control by manipulating the latent codes of GANs.} 
      In the first row, we adjust the facial attributes of a person's photo.
      In the second row, we use natural language sentences to alter the color and the shape of a generated flower. 
      In the last row, we edit keypoints of a generated anime character to modify head poses. 
      We use StyleGAN2~\cite{karras2020analyzing} to generate the images above.}
\end{figure}
\end{widetext}

\begin{abstract}
  \vspace{-5pt}
Generative adversarial networks (GANs) can generate high-quality images from sampled latent codes.
Recent works attempt to edit an image by manipulating its underlying latent code, but rarely go beyond the basic task of attribute adjustment.
We propose the first method that enables manipulation with multidimensional condition such as keypoints and captions.
Specifically, we design an algorithm that searches for a new latent code that satisfies the target condition based on the Surrogate Gradient Field (SGF) induced by an auxiliary mapping network.
For quantitative comparison, we propose a metric to evaluate the disentanglement of manipulation methods.
Thorough experimental analysis on the facial attribute adjustment task shows that our method outperforms state-of-the-art methods in disentanglement.
We further apply our method to tasks of various condition modalities to demonstrate that our method can alter complex image properties such as keypoints and captions.
\end{abstract}


\renewcommand{\thefootnote}{\arabic{footnote}}

\vspace{-7pt}
\section{Introduction}

Generative Adversarial Networks~\cite{goodfellow2014generative}, or GANs, are one of the most popular and effective methods for generating high fidelity images.
In the simplest form, the generator model creates a random image from a latent code sampled from the latent space.
To create an image that matches some target properties, however, we need a method to condition the generated image on such properties.
In other words, the method should be able to incorporate a piece of information, such as attributes, keypoints, or even an interpretation of the image in a natural language, into the generation of the image.
Intuitively, to condition the image, we can instead condition its latent code on the same information, in an attempt to generate an image that satisfies the target properties.

As an increasingly popular approach to image modification~\cite{abdal2019image2stylegan} and GAN interpretation~\cite{shen2020interpreting}, \textbf{latent space manipulation} is a type of approach that bases on varying the latent codes of images. 
The generator maps manipulated latent codes to images that hopefully match target properties.
To be specific, InterFaceGAN~\cite{shen2020interpreting} and GANSpace~\cite{harkonen2020ganspace} find meaningful directions in latent space, and vary latent codes along these directions to adjust the attributes of images.

Although existing methods explore the potential application of latent space manipulation, these methods still suffer from the following limitations.
To begin with, the disentanglement of manipulation can be limited. 
Adjustment of one attribute of an image is occasionally accompanied by some undesirable shifts in other attributes.
Moreover, existing methods are restricted to one-dimensional conditioning.
In other words, these methods excel in adjusting attributes such as smiling or not, female or male, each of which can be parameterized by a scalar condition.
However, these methods do not provide a general solution to complex modifications that condition on multidimensional information (\eg the pose of a human or the caption of an image).

We suggest that there is another line of latent space manipulation based on  optimization.
Using a generator and an image classifier, we can optimize the latent code for minimizing the difference between the properties of the current image and the target properties.
Empirically, this simple approach does not work as expected because both the classifier and the generator are highly non-convex deep neural networks.
As a result, the gradient field in the latent space may be misleading, and thus the optimization of a latent vector is often trapped in a local optimum.

To overcome the difficulty of the optimization-based approach, we propose a novel method for latent space manipulation. In our method, we train an auxiliary mapping network that induces a Surrogate Gradient Field (SGF).
We design an algorithm that uses SGF in search of a new latent code that satisfies a target condition.
For comparison with existing works, we design a metric that evaluates the  disentanglement of a manipulation method.
Based on the metric, we conduct thorough quantitative experiments and a user study to demonstrate that our method outperforms state-of-the-art methods in the  disentanglement of manipulation.
As the first work towards multidimensional conditioning with latent space manipulation, our method successfully modifies images utilizing keypoints and captions, illustrated with qualitative results.

To summarize our main contributions,
\begin{itemize}
  \setlength{\itemsep}{0cm}
  \item We propose the first latent space manipulation method of GANs that supports multidimensional conditioning. 
  \item We conduct quantitative experiments and a user study on the task of facial attribute adjustment to demonstrate that our method outperforms state-of-the-art methods in disentanglement.
  \item We apply our method to latent space manipulation using keypoints and captions, justifying our method as a unified approach for various modalities of conditioning.
\end{itemize}

\section{Related work}

\smallskip
\noindent\textbf{Generative Adversarial Networks.}
GAN~\cite{goodfellow2014generative} has shown great potential on generating photo-realistic images~\cite{radford2015unsupervised,karras2018progressive}.
It has been applied to a wide range of tasks including image editing~\cite{bau2019semantic, shen2020interpreting}, image translation~\cite{isola2016image, zhu2017unpaired} and super-resolution~\cite{ledig2017photo}.
Recent works have made tremendous progress on generating high-quality photo-realistic image~\cite{arjovsky2017wasserstein,gulrajani2017improved,brock2018large,karras2018progressive,karras2019style}.
Among the existing works on image generation, one of the most well-known works is StyleGAN~\cite{karras2019style} which introduces a stacked architecture that enables high-resolution image generation with fine-grained control.
Its recent follow-up work StyleGAN2~\cite{karras2020analyzing} further improved the generated image qualities and achieved state-of-the-art image synthesis results.
Our work greatly benefits from the progress of the GAN because we can apply our method to various GAN models.

\smallskip
\noindent\textbf{Manipulation on Latent Vector.}
Early GAN works~\cite{radford2015unsupervised} have already discovered that generated images can be semantically edited by applying vector arithmetic on the latent space.
Since vector arithmetic-based approach is straightforward and model agnostic, recent works continue to explore in this direction. 
Existing methods can be categorized into two classes: supervised methods \cite{shen2020interpreting,plumerault2019controlling,goetschalckx2019ganalyze} and
unsupervised methods~\cite{harkonen2020ganspace,voynov2020unsupervised}. 
Supervised methods use an extra classifier to label properties of generated images. Shen~\etal~\cite{shen2020interpreting} train a linear SVM on pairs of latent vectors and labels to find a decision hyperplane.
Latent vectors are then moved along the normal direction of the decision hyperplane for adjusting attributes.
For multiple attributes, their method can sacrifice performance for disentanglement by orthogonalizing each direction vector.
On the other hand, unsupervised methods directly find semantically meaningful directions by PCA~\cite{harkonen2020ganspace} or self-supervised learning~\cite{voynov2020unsupervised}.
Besides vector arithmetic-based approaches, some more recent works~ \cite{jahanian2019steerability,abdal2020styleflow} introduce non-linear transformations and generative modelings in the latent space to adjust multiple attributes simultaneously.

In contrast with existing methods, our approach utilizes a neural network to model complicated semantic relationships between latent vectors and corresponding predictions. We further extend the scope of conditions to a wider variety of vector representations. We show that our method achieve a higher degree of disentanglement compared with other methods.

\section{Method}
\renewcommand{\algorithmicrequire}{\textbf{Input:}} 
\renewcommand{\algorithmicensure}{\textbf{Output:}}

\subsection{Problem Definition}

Let $G \colon \mathcal{Z} \to \mathcal{X}$ be a pretrained GAN generator.
$\mathcal{Z} \subseteq \mathbb{R}^d$ is the $d$-dimensional latent space
\footnote{For StyleGAN, a latent vector $z$ is first sampled from a Gaussian distribution $\mathcal{N}(\mathbf{0}, \mathbf{I}_d)$ in Z-space, and a fully-connected neural network then transforms it into a new latent vector $w$ in W-space. In our formulation, $\mathcal{Z}$ can be either Z-space or W-space.}
, and $\mathcal{X}$ denotes the space of generated image.
The classifier network $C \colon \mathcal{X} \to \mathcal{C}$ predicts semantic properties $c \in \mathcal{C} \subseteq \mathbb{R}^{n_c}$ from a generated image $x \in \mathcal{X}$. 
Although $C$ can be as simple as a multi-label classifier, where $\mathbb{R}^{n_c}$ stands for the space of $n_c$ semantic attributes, the setting actually applies to any embedding in Euclidean space.
For example, keypoints detector with $n_p$ points on a 2D image can be regarded as an embedding to $\mathbb{R}^{2 n_p}$.

Define $\Phi(z) \coloneqq C(G(z))$ for convenience. 
Suppose we have a latent vector $z_0 \in \mathcal{Z}$, its corresponding properties $c_0 = \Phi(z_0)$ and target properties $c_1$. Our goal is to find $z_1 \in \mathcal{Z}$ such that $\Phi(z_1) = c_1$. 

\subsection{Learning the Auxiliary Mapping}

A powerful generator such as StyleGAN2~\cite{karras2020analyzing} may easily generate infinite images that match the properties $c_1$.
We would like to attain the desired properties $c_1$ with minimal unwanted modification to the image.
Intuitively, in $\mathcal{Z}$ space, $z_0$ can be slightly perturbed to get a $z_1$ that is sufficiently close to  $z_0$. 
Empirically, the gradient field of $\Phi$ is not suitable for perturbing $z_0$, so we seek to replace it with a new gradient field. 

As a preparation, we introduce an auxiliary mapping $F:\mathcal{Z} \times \mathcal{C} \to \mathcal{Z}$ satisfying
\begin{equation} \label{eq:F}
F(z, \Phi(z)) = z, \forall z \in \mathcal{Z}
\end{equation}
In our implementation, $F$ is a multi-layer neural network, and trained using a simple reconstruction loss.
Inspired by Behrmann \etal~\cite{behrmann2019invertible}, we use spectral normalization~\cite{miyato2019spectral} in $F$ so that its Lipschitz constant $\mathrm{Lip}(F) < 1$.
As a result, the operator norm of its Jacobian is less than 1~\cite{horn_johnson_2012}. 
Furthermore, for any eigenvalue $\lambda_F$ of the Jacobian of $F$ and the corresponding unit eigenvector $x_F$, we have $
\left\|\lambda_F x_F\right\| 
= \left\|\frac{\partial F(z, c)}{\partial z} x_F\right\| 
\leq \left\|\frac{\partial F(z, c)}{\partial z}\right\|_{\mathrm{op}} 
< 1$, where $ \left\|\cdot \right\|_{\mathrm{op}}$ denotes operator norm. 
Therefore, the spectral radius of the Jacobian of $F$ satisfies

\begin{equation} \label{eq:F_jacobian}
\rho\left(\frac{\partial F(z, c)}{\partial z}\right) \leq \left\|\frac{\partial F(z, c)}{\partial z} \right\|_{\mathrm{op}} < 1
\end{equation}

\begin{figure}[!t]
\begin{center}
   \includegraphics[width=0.75\linewidth]{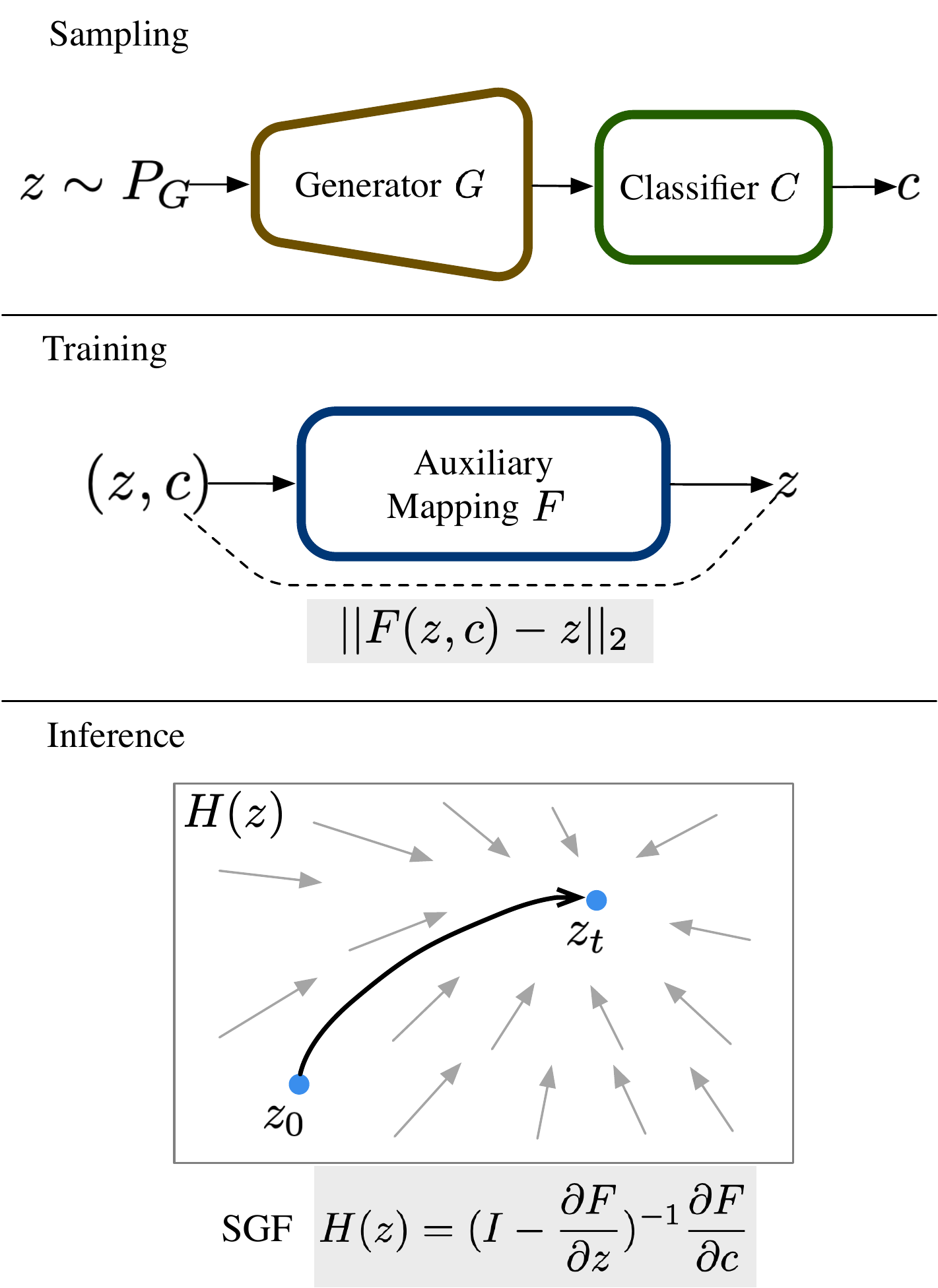}
\end{center}
\vspace{-5pt}
   \caption{\textbf {Overview of our method.} 
   $P_G$ denotes the distribution of latent vectors in a latent space, which can be either Z-space or W-space in the case of StyleGAN. 
   We sample $(z,c)$ pairs, and train the auxiliary mapping $F$ using MSE loss. 
   The surrogate gradient field $H$ navigates the latent vector to the target in the inference stage.}
\label{fig:model-training}
\vspace{-5pt}
\end{figure}

Figure~\ref{fig:model-training} shows the training pipeline of $F$.
\subsection{Manipulation with Surrogate Gradient Field}
 
To formalize the perturbation of $z_0$, we define a path $z(t), t \in [0, 1]$ in the latent space that starts from $z_0$ and ends at $z_1$, i.e. $z(0) = z_0$ and $z(1) = z_1$.
Here we make several assumptions about path $z(t)$.
\begin{enumerate}
\setlength{\itemsep}{0cm}
\item The generator is capable of generating an image that match the desired properties: $$
\exists\ z_1 \in \mathcal{Z} \quad s.t. \quad  \Phi(z_1) = c_1
$$

\item While traversing the path, the properties $\Phi(z(t))$ of the generated image changes at a constant rate, i.e.
\begin{equation} \label{eq:const_rate}
\frac{\mathrm{d}\Phi(z(t))}{\mathrm{d} t} = c_1 - c_0 
\end{equation}

\item 
$\forall\ z \in \mathcal{Z}$,
\begin{equation} 
\frac{\partial F(z, \Phi(z))}{\partial c} \neq 0 
\end{equation}
\end{enumerate}

The assumptions above suggests that 1. our task is well-posed, 2. path $z(t)$ is a smooth interpolation between the original properties and the target properties, and 3. $F$ is not a trivial mapping that just map any $(z, c)$ pair to $z$. 

Now we derive the surrogate gradient field of $\Phi$. Using Eq.~(\ref{eq:F}) of auxiliary mapping $F$, we can rewrite the path as
\begin{equation}
z(t) = F(z(t), \Phi(z(t)))
\end{equation}

Take time derivatives on both sides, we have
\begin{equation*}
\begin{split}
&\quad\frac{\mathrm{d} z(t)}{\mathrm{d} t} = \frac{\mathrm{d}F(z(t), \Phi(z(t)))}{\mathrm{d}t} \\
&= \frac{\partial F(z(t), \Phi(z(t)))}{\partial z} \frac{\mathrm{d} z(t)}{\mathrm{d} t} + 
\frac{\partial F(z(t), \Phi(z(t)))}{\partial c} \frac{\mathrm{d}  \Phi(z(t))}{\mathrm{d} t} \\
&= \frac{\partial F(z(t), \Phi(z(t)))}{\partial z} \frac{\mathrm{d} z(t)}{\mathrm{d} t} + 
\frac{\partial F(z(t), \Phi(z(t)))}{\partial c}(c_1 - c_0) 
\end{split}
\end{equation*}

We plug in assumption 2 in the last step. Organize $\frac{\mathrm{d} z(t)}{\mathrm{d} t}$ to the left hand side and rearrange the last equation, we have
 $\frac{\mathrm{d} z(t)}{\mathrm{d} t} = 
\left( \mathbf{I}  - \frac{\partial F(z(t), \Phi(z(t)))}{\partial z}\right)^{-1}
\frac{\partial F(z(t), \Phi(z(t)))}{\partial c}
(c_1 - c_0)$, the invertibility implied by Eq.~(\ref{eq:F_jacobian})~\cite{horn_johnson_2012}.

Define surrogate gradient field $H$ as
\begin{multline}\label{eq:5}
H(z) \!\!\coloneqq \!\!
\left( \!\mathbf{I}  \!-\! \frac{\partial F(z, \Phi(z))}{\partial z}\right)^{\!\!-1} 
\!\! \frac{\partial F(z, \Phi(z))}{\partial c} 
(c_1 - c_0) \!\!
\end{multline}

Note that $H(z) \neq 0, \forall z \in \mathcal{Z}$ because of Eq.~(\ref{eq:F_jacobian}) and assumption 3.
We arrive at our ordinary differential equation,
\begin{equation}\label{eq:ode}
\begin{split}
\begin{cases}
\frac{\mathrm{d} z(t)}{\mathrm{d} t} \!\! &= \ H(z(t)),\ t \in [0,1] \\
z(0) \!\! &= \ z_0
\end{cases}
\end{split}
\end{equation}

\begin{figure}[!t]
\vspace{-7pt}
  \begin{algorithm}[H]
    \caption{Manipulating GAN with surrogate gradient field}\label{alg:1}
    \begin{algorithmic}
      \Require Generator $G$, Classifier $C$, auxiliary mapping $F$, order of the series expansion $m$, iteration number $n$, initial latent vector $z_0$, target attributes $c_1$, step size $\lambda$
      \State $c_0 \leftarrow C(G(z_0))$
      \State $\delta_c \leftarrow \lambda (c_1 - c_0)$
      \State $c^{(0)} \leftarrow c_0$
      \For{$i=1,\cdots,n$}
        \State $\delta_z^{(0)} \leftarrow \frac{\partial F}{\partial c}(z^{(i-1)}, c^{(i-1)}) \delta_c$
        \State $\delta_z \leftarrow \delta_z^{(0)} $
        \For{$j=1,\cdots,m$}
            \State $\delta_z^{(j)} \leftarrow \frac{\partial F}{\partial z}(z^{(i-1)}, c^{(i-1)}) \delta_z^{(j - 1)}$
            \State $\delta_z \leftarrow \delta_z + \delta_z^{(j)} $
        \EndFor
        \State $z^{(i)} \leftarrow z^{(i-1)} + \delta_z$
        \State $c^{(i)} \leftarrow C(G(z^{(i)}))$
        \If{$c^{(i)}$ close to $c_1$ }
            \State \Return $z^{(i)}$
        \EndIf
      \EndFor
      \State \Return $z^{(n)}$
    \end{algorithmic}
  \end{algorithm}
  \vspace{-5pt}
  \caption{
\textbf{Pseudocode of our manipulation algorithm.}
The outer loop is a simple forward Euler ODE solver, which computes the movement $\delta_z$, and accumulate to the current latent vector $z^{(i)}$.
The classifier predicts the properties of image at each time step to determine when to stop.
The inner loop approximates the matrix inversion term in Eq.~\eqref{eq:5} using the Neumann series.}
\end{figure}

\subsection{Numerical Solution of the ODE}

To compute our goal $z(1)$, we solve the initial value problem (Eq.~(\ref{eq:ode})) using a numerical ordinary differential equation solver.
Nevertheless, it is time consuming and potentially numerically unstable to calculate the Jacobian of $F$ and the matrix inversion when evaluating $H(z)$ (Eq.~\eqref{eq:5}).
Instead, we apply Neumann series expansion~\cite{horn_johnson_2012} to approximate the matrix inversion.
For a matrix $\mathbf{X}$ that satisfies $\rho(\mathbf{X}) < 1$, the following expansion converges
$$
(\mathbf{I} - \mathbf{X})^{-1} = \mathbf{I} + \mathbf{X} + \mathbf{X}^2 + \ldots
$$

Another obstacle to numerical computation is that, in reality, the path may deviates from the assumption 2. To be specific, at step $i$ with a step size of $\lambda$, 
$\Phi(z( i \lambda))$ does not precisely equals 
$\Phi(z((i - 1)\lambda)) + \lambda(c_1 - c_0)$. 
Two source of error leads to the problem: one from the numerical solver, and another from not having a perfect $F$ which has $F(z, \Phi(z)) = z$ exactly everywhere.
To overcome this difficulty, in practice we fix the step size $\lambda$ but do not necessarily stop the iteration process at step $1/\lambda$. 
The algorithm checks the properties $c_i = \Phi(z(i \lambda))$ at each step, and stops only when $c_i$ is sufficiently close to the target $c_1$, unless it reaches the maximum step number.
Algorithm~\ref{alg:1} shows the summary of the manipulation procedure.

\section{Experiments}
\subsection{Compared Methods}

We compare the proposed method \ourmethod with two state-of-the-art latent space manipulation methods: 
\textbf{InterfaceGAN}  \cite{shen2020interpreting} \footnote{https://github.com/genforce/interfacegan} and
\textbf{GANSpace} \cite{harkonen2020ganspace} \footnote{https://github.com/harskish/ganspace}.
All compared methods are tested using the official code release.

\smallskip
\noindent\textbf{InterfaceGAN.}
We retrain the InterfaceGAN model for each control attribute.
Since InterfaceGAN can only learn one binary attribute at once, we train on each attribute independently with the same training data of our \ourmethod strictly following the training setting in the paper.

\smallskip
\noindent\textbf{GANSpace.}
For GANSpace, we use the pre-selected control vectors released in its official code and only apply changes to the recommended StyleGAN2 layers.

\subsection{Generator Models and Datasets}

Choosing different combinations of the latent space $\mathcal{Z}$ and condition space $\mathcal{C}$, we set up four distinct settings for latent space manipulation to demonstrate that our method can control different generator models under various types of conditions.

For the generator, we test StyleGAN2~\cite{karras2020analyzing} and ProgressiveGAN~\cite{karras2018progressive}. 
StyleGAN2 experiments are conducted on W-space, while ProgressiveGAN experiments are conducted on Z-space.
To further demonstrate that our method can accept various types of conditions besides image attributes, we conduct experiments on two other representative properties (\ie, keypoints and image captions).
We only show the results of our \ourmethod method for keypoints and image captions, since other methods are not able to utilize these conditions.

\smallskip
\noindent\textbf{FFHQ-Attributes.}
We adopt a pretrained FFHQ StyleGAN2~\cite{karras2020analyzing} as the generator for experiments on facial attributes editing.
For the classifier, we fine-tune a pretrained SEResNet50~\cite{hu2018squeeze} model from VGGFaces2~\cite{cao2018vggface2} dataset.
We construct the training data for the classifier model by labeling $100K$ randomly sampled images with the Azure Face API~\footnote{https://azure.microsoft.com/en-us/services/cognitive-services/face/}, and combine them with labeled faces from the CelebA~\cite{liu2015deep} dataset. 
With duplicate labels removed, the final classifier can predict $48$ facial attributes.
Among them, we select four representative attributes, which includes both highly entangled attributes (``gender'' and ``bald'') and less entangled ones (``smile'' and ``black hair''), for quantitative comparisons and the user study.

\smallskip
\noindent\textbf{CelebAHQ-Attributes.}
To compare the performance on models other than StyleGAN, we also test a ProgressiveGAN~\cite{karras2018progressive} pretrained on the CelebAHQ dataset. 
We use the same facial attributes classifier as the FFHQ-Attributes in this experiment.

\smallskip
\noindent\textbf{Anime-KeypointsAttr.} 
We follow~\cite{jin2017towards, saito2015illustration2vec} to build a high-quality Japanese anime-face dataset and train a StycleGAN2 on it.
We base on the animeface-2009~\footnote{https://github.com/nagadomi/animeface-2009} and illustration2vec~\cite{saito2015illustration2vec} to create facial landmarks keypoints and image attributes as the conditions for manipulation.

\smallskip
\noindent\textbf{Flowers-Caption.}
Previous works have shown great success on training GANs conditioned on text captions~\cite{zhang2016stackgan}. 
However, to our best knowledge, \ourmethod is the first method that can utilize text captions to conditionally manipulate latent vectors of a pretrained GAN. 
Our experiment is based on a pretrained image generator model~\cite{zhao2020differentiable} on Oxford-102 Flowers dataset~\cite{nilsback2008automated}.
The image caption generator is an attention-based caption model~\cite{xu2015show} trained on flower caption dataset~\cite{reed2016generative}.
To fit our pipeline for latent space manipulation, we use the sentence transformer~\cite{reimers2019sentence} to encode generated captions into vectors.

\begin{figure}[t]
\begin{center}
  \centering
  \includegraphics[width=.48\textwidth]{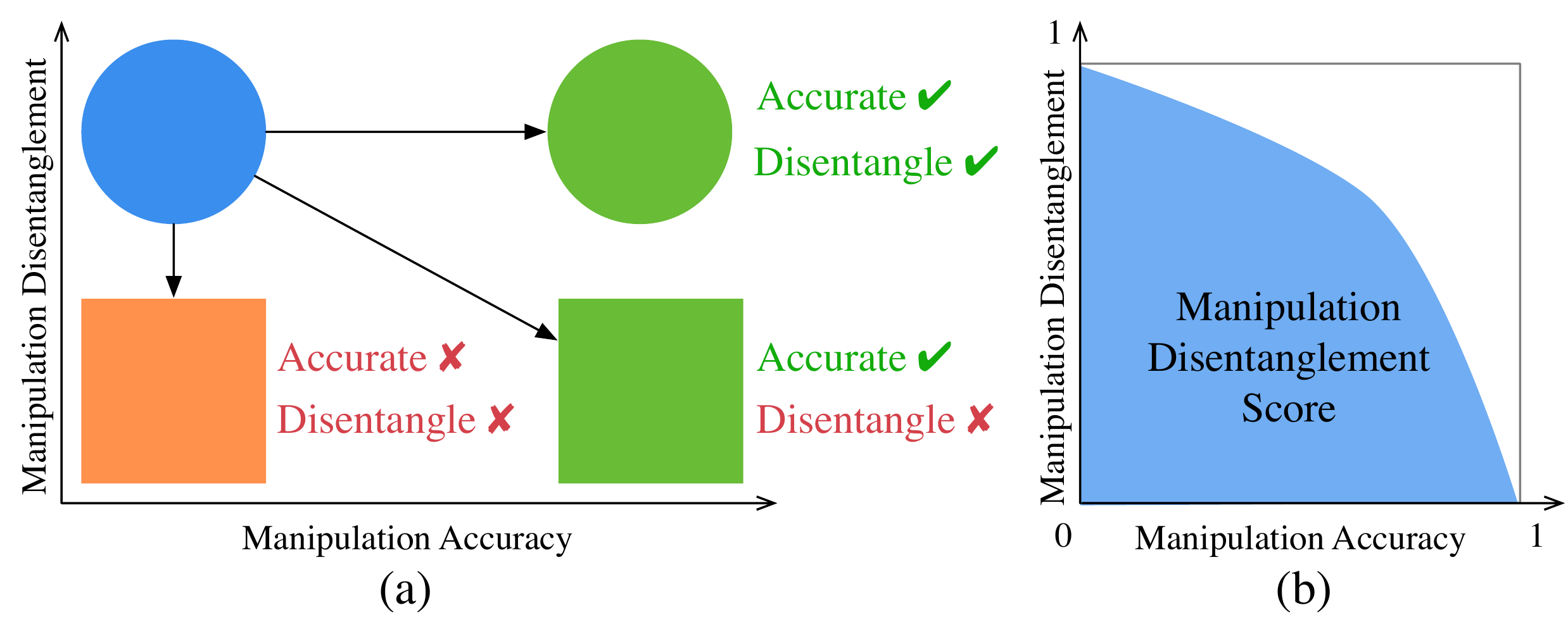}
  \caption{
  \textbf{Illustration of the Manipulation Disentanglement Score (MDS).} 
  (a) Manipulating the color of a blue circle to green while keeping the shape unchanged. 
  (b) MDS is defined as the AUC of Manipulation Disentanglement Curve (MDC). 
  }\label{fig:metric-example}
  \vspace{-15pt}
\end{center}
\end{figure}

\subsection{Implementation Details}

The auxiliary mapping $F$ is implemented with an $N$-layer MLP combined with AdaIN~\cite{huang2017arbitrary}.
We also apply spectral normalization~\cite{miyato2019spectral} to all fully-connected layers.
We describe the detail of network architectures in the supplementary material.
For Z-space experiments, we set $N=6$. While, for W-space experiments, we observe that $F$ can easily degenerate to a trivial mapping by ignoring conditions $c$ when $N=6$. To prevent the degeneration, we increase $N$ to $15$ for all W-space experiments.
For each experiment, we sample $200k$ pairs of latent vectors and corresponding conditions to build the training dataset of $F$. 
We apply a truncation rate of $0.8$ to all StyleGAN2 samples.
We train $F$ for $500k$ iterations with a batch size of $8$ using Adam optimizer~\cite{kingma2015adam} with learning rate of $0.0002$.
For the manipulation, we apply Algorithm~\ref{alg:1} with order $m=1$ and step size $\lambda = 0.2$ as default.

\subsection{Evaluation Metrics}

It is difficult to designing comprehensive quantitative metrics for measuring the disentanglement of latent space manipulation methods, which often use model-specific hyper-parameters to control the editing strength.
For example, Figure~\ref{fig:ffhq-attr}(b) shows manipulation results of ``gender'' from different methods under different editing strength.
Shen~\etal~\cite{shen2020interpreting} use the number of prediction changes to measure disentanglement among different attributes. 
However, comparing only the final results of image manipulation algorithms can be unfair.
When editing strength increases, some methods tend to over-modify the image, i.e. introducing unwanted modification.
Therefore, for comprehensive measurement of disentanglement, it is necessary to design an editing strength-agnostic metric.

\vspace{-5pt}
\subsubsection{Manipulation Disentanglement Score} \label{sec:mds}

For a given manipulation goal, a trade-off between accuracy and disentanglement often exists.
Figure~\ref{fig:metric-example}(a) illustrates the possible ways to change a blue circle to a green one.
For a both accurate and disentangled manipulation, the color becomes green while the shape keeps round.
An example of accurate but entangled manipulation would be changing the shape to a square when the color turns green.

By gradually increases manipulation strength and calculate the accuracy and disentanglement measure at each point, we can plot these points on the accuracy-disentanglement plane to attain a \textbf{Manipulation Disentanglement Curve} (MDC).
As Figure~\ref{fig:metric-example}(a) suggests, a method with an MDC closer to $y=1$ indicates overall better disentanglement.
In this way, we can compare the MDCs with each other in different methods.

In reminiscence of ROC curve, we define \textbf{Manipulation Disentanglement Score} (MDS) as the Area under Curve (AUC) of an MDC, illustrated in Figure~\ref{fig:metric-example}(b).
A method with a higher MDS suggests that it has a higher degree of disentanglement for the given manipulation.

For an experiment of attributes manipulation with $N$ samples, suppose we can infer the scores of $M$ attributes in total from an image.
We consider an attribute is changed if the score changes more than $0.5$ during the manipulation.
Suppose there are $N_s$ sample which successfully have their attributes changed to the target attributes. 
The manipulation accuracy is then the success rate $N_s / N$.
For sample $i$, if $n_i$ attributes \emph{other than} the target attribute have changed, we can use $ \frac{1}{N} \sum_{i = 1}^{N} (1 - \frac{n_i}{M-1})$ as the manipulation disentanglement.
An alternative way to define manipulation disentanglement is using image similarity, however, we found it less sensitive to subtle changes like added beards compared to the image attribute classifier we use.
In our experiments on facial attributes manipulation, we evaluate $N=100$ samples for each attribute, and $M=48$.
We inverse the direction of manipulation for samples that already match the target attribute so that we can calculate manipulation accuracy for every sample.

\vspace{-5pt}
\subsubsection{User Study}

In addition to quantitative comparison on MDS, we conduct a user study in the FFHQ facial attributes experiments to further evaluate the disentanglement of methods.
For each question of the user study, a user would see a source image and manipulation results from both our \ourmethod and the InterfaceGAN. 
The user is then asked to choose a result that has best changed the source image to match a target attribute while keeping other features unchanged.
We use $10$ random generated images and $10$ photos projected to the latent space of GAN~\cite{karras2020analyzing}.
In total, $20$ participants have made $400$ preference choices.

\subsection{Comparisons on FFHQ-Attributes} \label{sec:comp-attr}

\begin{figure*}[t]
\begin{center}
   \centering
  \includegraphics[width=0.90\textwidth]{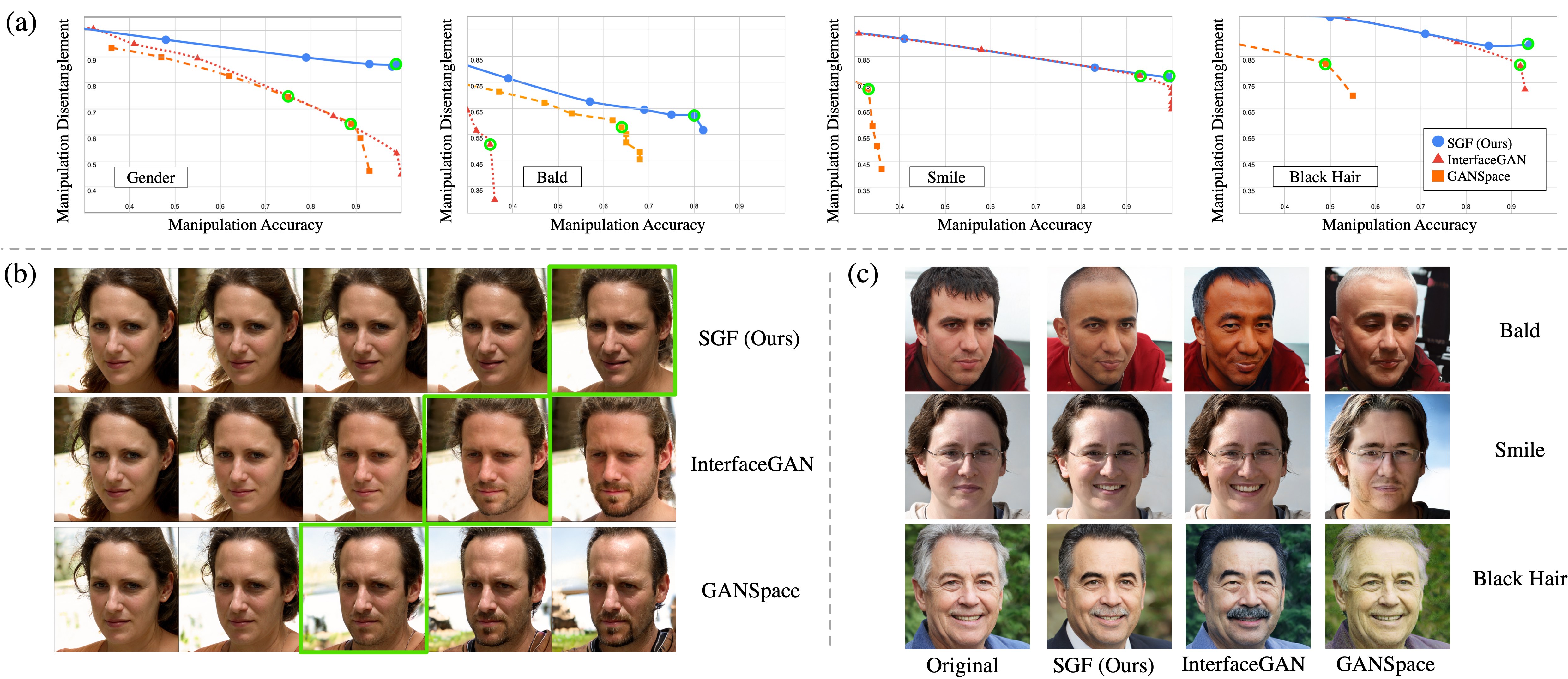}
   \vspace{-2pt}
  \caption{
  \textbf{Comparison of facial attribute editing in the FFHQ-Attributes.}
  (a) The MDCs of methods for each attribute. The point highlighted with a green circle has highest harmonic mean of accuracy and disentanglement along the curve.
  (b) ``gender'' manipulation results of different methods. Green boxes mark the results that use the highlighted hyper-parameters.
  (c) Manipulation of other attributes. We use the highlighted hyper-parameters of each method.
  }
  \label{fig:ffhq-attr}
  \vspace{-15pt}
\end{center}
\end{figure*}

\begin{table}[t]
\caption{
    \textbf{MDS comparison on facial attribute editing on FFHQ-Attributes and CelebaHQ-Attributes.}
    Our SGF method shows the best overall score in attribute editing experiments on both FFHQ and CelebaHQ datasets, and significantly outperforms the compared methods on attributes that tend to be entangled (\eg ``gender'' and ``bald'').
\vspace{-5pt}
} 
\begin{center}
\resizebox{0.85\columnwidth}{!}{
\begin{tabular}{lcccccc}
\hline
 & \multicolumn{4}{c}{\bf{MDS on FFHQ-Attributes}}
 & \multicolumn{1}{c}{~} \\
\bf{Method} & \bf{Gender}  &  \bf{Bald} & \bf{Smile} & \bf{Black Hair} &  \bf{Overall}\\
\hline  
GANSpace & 0.841 &  0.491 & 0.248 & 0.543 & 0.531 \\ 
InterfaceGAN & 0.808 &  0.254 & \bf0.883 & 0.938 & 0.721 \\ 
\ourmethod (Ours) & \bf0.919 & \bf0.590 & \bf0.884 & \bf0.955 & \bf0.837 \\
\hline
\hline
 & \multicolumn{4}{c}{\bf{MDS on CelebAHQ-Attributes}}
 & \multicolumn{1}{c}{~} \\
\bf{Method} & \bf{Gender} &  \bf{Bald} & \bf{Smile} & \bf{Black Hair} &  \bf{Overall}\\
\hline  
InterfaceGAN & 0.876  & 0.442 & 0.856 & 0.876 & 0.758 \\ 
\ourmethod (Ours) & \bf0.912 & \bf0.799 & \bf0.896 & \bf0.897 & \bf0.876 \\
\hline
\end{tabular}
}
\end{center}
\vspace{-10pt}
\label{tab:ffhq-mds}
\end{table}

Experiments on attributes manipulation compare SGF to the baseline models in the perspectives of manipulation disentanglement and accuracy defined in Sec.~\ref{sec:mds}.
In Figure~\ref{fig:ffhq-attr}(a), we plot the Manipulation Disentanglement Curves (MDCs) for our proposed \ourmethod with state-of-the-art methods on four facial attribute editing settings.
Our method has shown a better or comparable disentanglement degree compared with other methods.

From the MDC of ``gender'' in baseline methods, we observe a sacrifice of manipulation disentanglement for high accuracy, which suggests that high manipulation strength in baseline methods introduces changes in non-target attributes .
Figure~\ref{fig:ffhq-attr}(b) qualitatively compare the results of editing ``gender'' attribute.
Our method changes ``gender'' without side effects such as adding beards. 
In contrast, both the InterfaceGAN and GANSpace add non-target properties to the final results when manipulation strength is high.
We make the same observation on the ``gender'' MDC in Figure~\ref{fig:ffhq-attr}(a): as accuracy increases with the manipulation strength, the disentanglement degree of all methods except SGF drops significantly.
This suggests that while accuracy of baseline methods comes at the price of entanglement, our method is able to achieve high accuracy and disentanglement at the same time.

In Figure~\ref{fig:ffhq-attr}(c), we qualitatively compare SGF with InterfaceGAN and GANSpace on editing other attributes.
For each method and attribute, we use the hyper-parameters in settings highlighted with green circles in Figure~\ref{fig:ffhq-attr}(a).
For each highlighted setting, the harmonic mean of accuracy and disentanglement reach the peak on the curve.
while editing the target attribute, SGF consistently changes the least number of other properties.
InterfaceGAN achieves similar disentanglement in ``smile'', while showing inferior results in both ``bald'' and ``black hair''.
GANSpace shows inferior results in all settings.

We calculate the AUC for each method and attribute in Figure~\ref{fig:ffhq-attr}(a) as the MDS in Table~\ref{tab:ffhq-mds}.
We find some attributes tend to correlate with others, \eg ``bald'' often correlates with ``gender'' (Figure~\ref{fig:ffhq-attr}(b)).
For experiments of such attributes, our proposed method significantly outperforms others.
For editing relatively less entangled attributes, \eg ``smile'' and ``black hair'', our method has comparable results with InterfaceGAN and outperforms GANSpace.
These results also align with the visual perception for each image in Figure~\ref{fig:ffhq-attr}(b) and (c).
The overall score shows our method can generally achieve better disentanglement with high manipulation accuracy than InterfaceGAN and GANSpace.
As GANSpace shows inferior overall performance, we only compare our method with InterfaceGAN in the following experiments.

In our user study for comparison of SGF with InterfaceGAN,
61\% of the total queries ($244$ queries of the total $400$ queries) judge our method has a higher degree of disentanglement.
Combining the results with the experiments on MDS, we conclude that our method is able to edit attributes with less entanglement compared with other methods.

\begin{figure*}[t!]
\begin{center}
  \centering
  \includegraphics[width=0.88\textwidth]{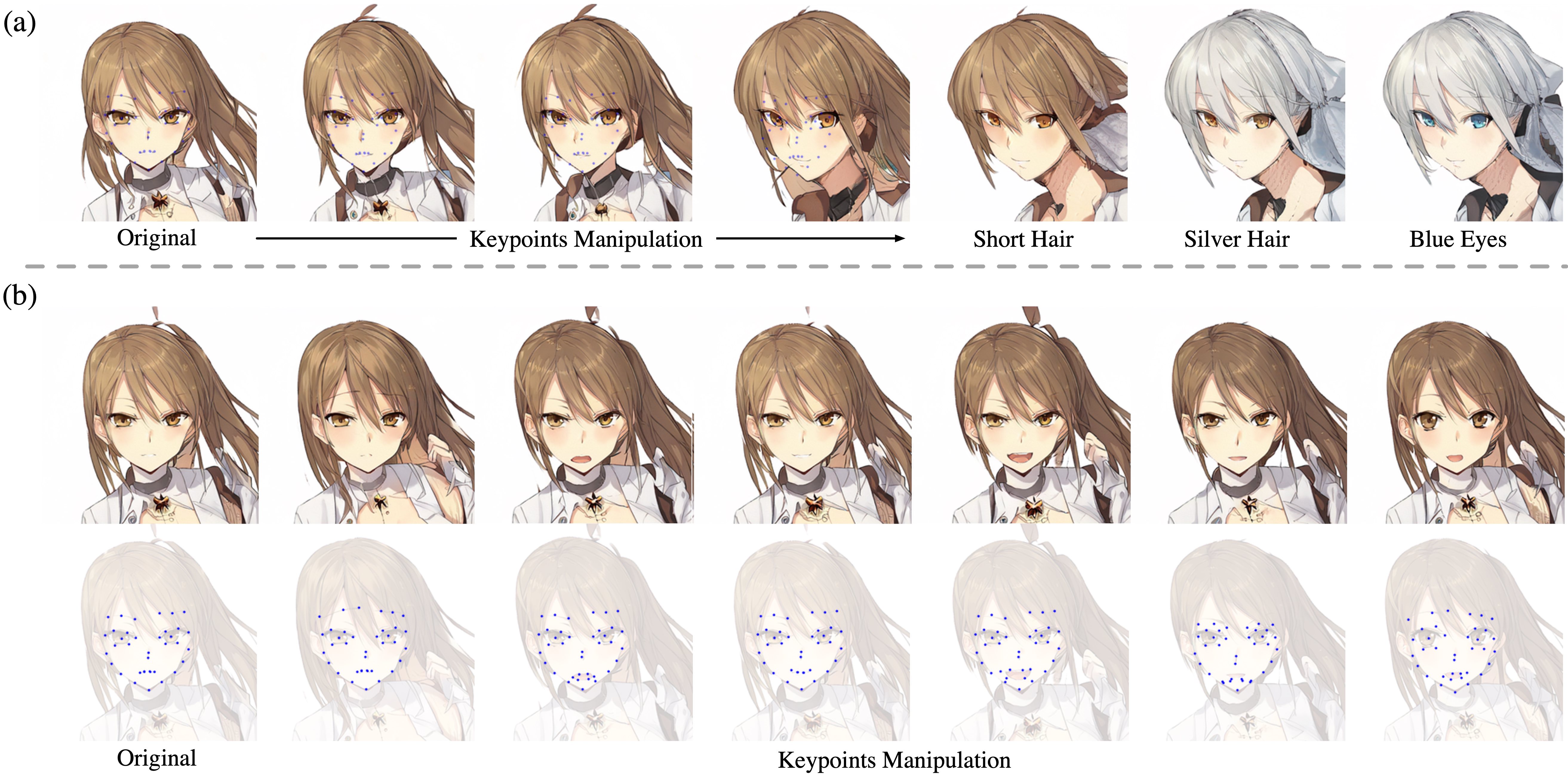}
  \caption{\textbf{Manipulation in Anime-KeypointsAttr dataset.}
    (a) Sequential editing by keypoints (column 1 to 4) and attributes (columns 5 to 7). Target keypoints are shown as blue dots.
    (b) Keypoints manipulation for expression control, the second row shows the corresponding target keypoint conditions.
  }\label{fig:anime-keypoints}
\vspace{-15pt}
\end{center}
\end{figure*}

\begin{figure*}[t]
\begin{center}
   \centering
  \includegraphics[width=0.95\textwidth]{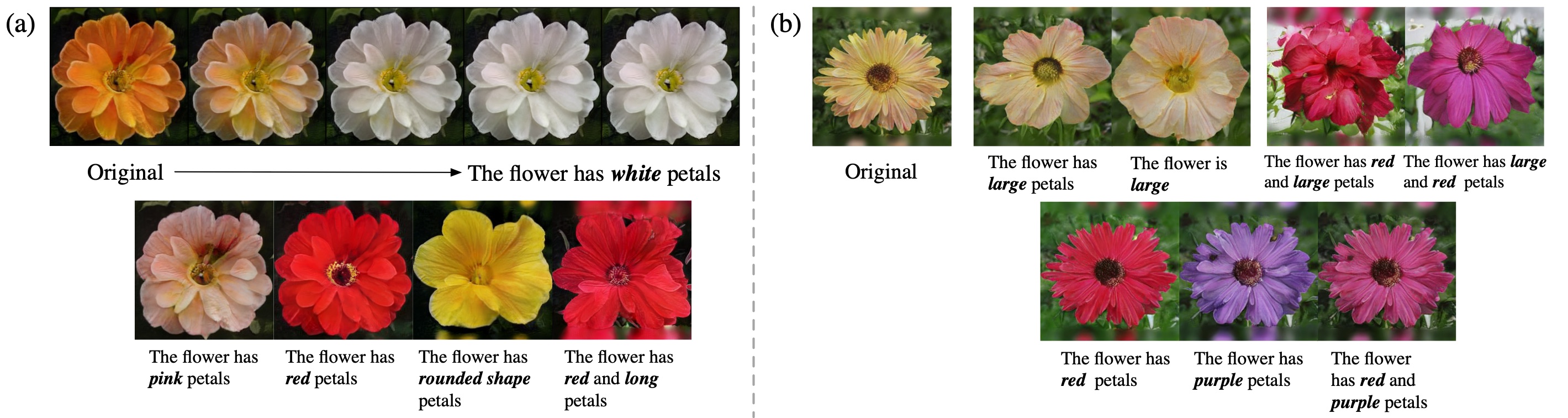}
  \caption{
  \textbf{Manipulation by caption in Flowers-Caption dataset.}
  (a) Latent space manipulation results on Flowers-Caption using different target captions.
  (b) Manipulation of Flowers with different caption compositions.
  }\label{fig:flowers-caption}
\vspace{-15pt}
\end{center}
\end{figure*}

\subsection{Comparison on CelebAHQ-Attributes}

The MDS of CelebAHQ-Attributes data are shown in Table~\ref{tab:ffhq-mds}.
Despite using a different GAN model, our SGF still outperforms InterfaceGAN with a similar margin in each attribute in FFHQ data.
These results indicate that our method can be applied to different GAN models while maintaining similar performance gains compared to InterfaceGAN.

\subsection{Manipulation on Anime-KeypointsAttr} \label{sec:keyp-caption}

Extending the control conditions to keypoints-attributes, we demonstrate that \ourmethod can use keypoints and attributes to jointly control anime faces.
Figure~\ref{fig:anime-keypoints}(a) shows the sequential editing results of head poses and facial attributes.
Our model edit images in a stable and disentangled manner throughout the manipulation process of both keypoints and attributes.

By fine-tuning each facial keypoint, we can add precise facial expression control to anime characters.
As shown in Figure~\ref{fig:anime-keypoints}(b), moving the eyebrows changes the overall expression from natural to sad in the second column. In other columns, we controls the mouth and eyes to change the character's expressions (e.g., angry or happy). 

\subsection{Manipulation on Flowers-Caption}

To further explore the potential of multi-dimensional control,
we use natural language as control conditions with the help of sentence embedding.
Figure~\ref{fig:flowers-caption} (a) shows that our method can manipulate the color and the shape of generated flowers according to the given target captions.

Figure~\ref{fig:flowers-caption}(b) shows manipulation results with different caption compositions.
The first row compares the results using captions with similar meanings.
While ``large and red'' and ``red and large'' produce completely different flowers, both results match the target caption.
The images in the second row show the results of color mixing.
The manipulation result of ``red and purple'' is a flower with purplish-red petals.
From caption compositions experiments, we suggest that our method can leverage the power of sentence embedding to manipulate latent codes. 

\subsection{Limitations and Discussions} \label{sec:limit}

\begin{figure}[!t]
\begin{center}
  \includegraphics[width=1\linewidth]{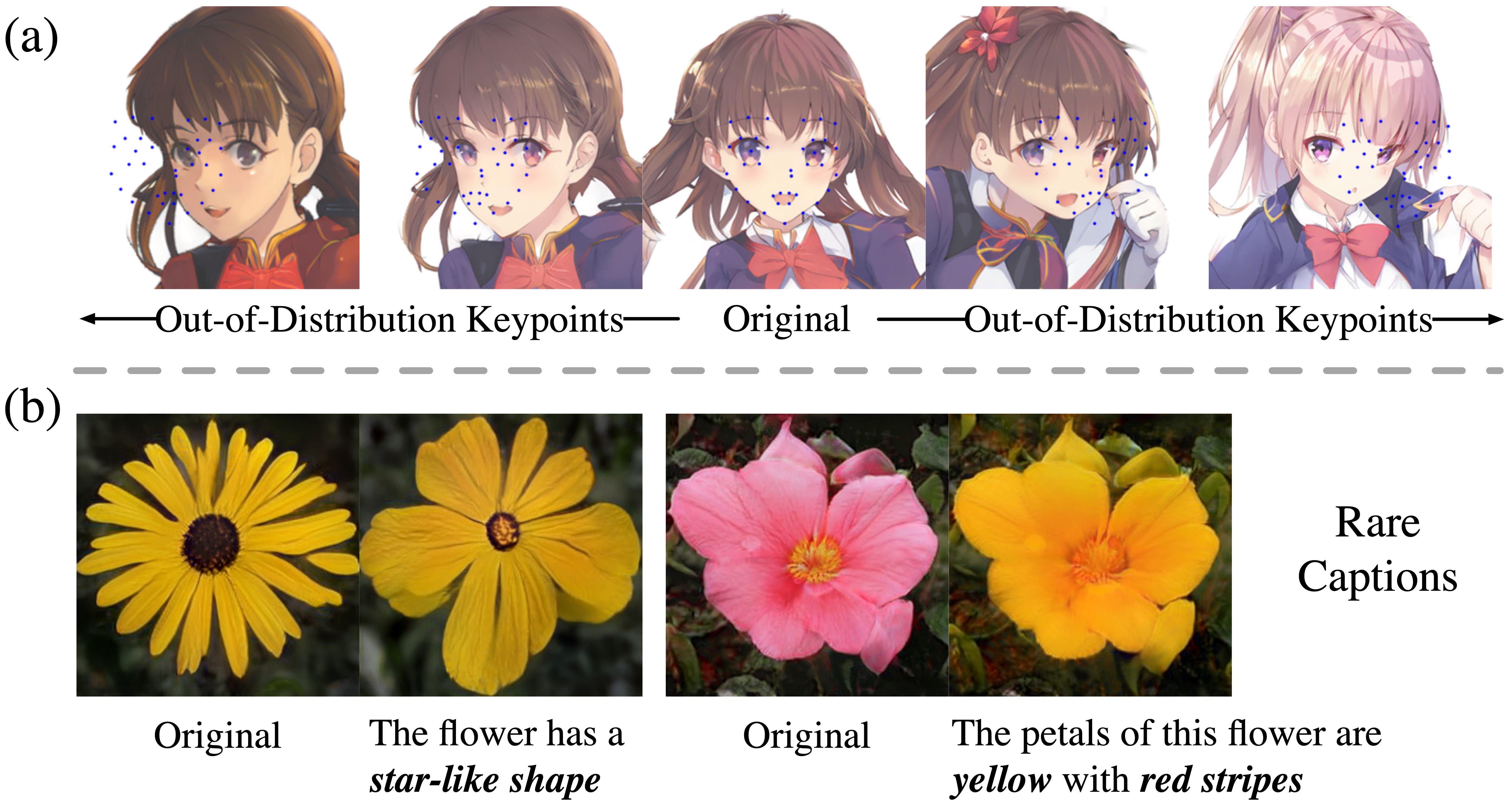}
\end{center}
\vspace{-10pt}
  \caption{\textbf{Typical failure cases of our method.}
  Refer to Section~\ref{sec:limit} for details.
}
\vspace{-15pt}
\label{fig:limit}
\end{figure}

Some limitations exists for SGF despite the compelling experimental results.
Figure~\ref{fig:limit} shows typical failure cases of SGF.
To begin with, \ourmethod does not cover the case where target condition is out of the training data distribution.
For an anime image generator trained on aligned face images, faces with unaligned keypoints are out of the generation scope.
Therefore, for the results of head yaw modification using keypoints in Anime-KeypointsAttr dataset (Figure~\ref{fig:limit}(a)), 
the edited faces do not exactly match the given target keypoints.
If the target condition is relatively near the generation scope, our method tends to stop at a point with a similar condition.
However, an extremely out-of-distribution target condition may lead to side effects including style and color changes
(the leftmost and rightmost images in Figure~\ref{fig:limit}(a)).
In addition, there are cases where our model fails to capture conditions that rarely appear.
For example, \ourmethod failed to edit flowers in Figure~\ref{fig:limit}(b) because both captions are uncommon in the training dataset.
We suggest that building a high-quality dataset with diversity and balanced distribution of condition may be the key to overcome the above limitations.

\vspace{-3pt}
\section{Conclusions}
\vspace{-2pt}
We proposed a unified approach for latent space manipulation on various condition modalities, showed a higher degree of disentanglement in facial attributes editing and able to use facial landmarks as well as natural languages to edit an image.
The multi-dimensions control has the potential application to a wide variety of settings and we hope this method will provide interesting avenues for future work.

\smallskip
\noindent\textbf{Acknowledgments.}
We thank Yingtao Tian for helpful discussions and all reviewers for valuable comments.

\appendix
\section*{\Large{Appendices}}
\vspace{10pt}

\vspace{-7pt}
\section{Simple Optimization on Latent Code}

In this section, we demonstrate that simple latent code optimization fails to alter the attributes of a given image. 
As the most straightforward method to manipulate the latent space, latent code optimization first calculates the difference between the current attributes and some desired attributes, and then backpropagates the error to the latent vector. 
As we update the latent vector in each step to minimize the difference, we expect that the optimized result should possess the desired attributes.

Specifically, we use the following setting for latent code optimization.
Given original latent vector $z_0$, its corresponding attributes $c_0$ and the target attributes $c_1$.
We set the initial value $z = z_0$, and optimize $z$ via back-propagation to get the result $z_1 = \argmin_{z} ||C(G(z)) - c_1||_2$.
We use the Adam~\cite{kingma2015adam} optimizer with learning rate set to $0.0002$.

\begin{figure}[!t]
\begin{center}
  \includegraphics[width=0.8\linewidth]{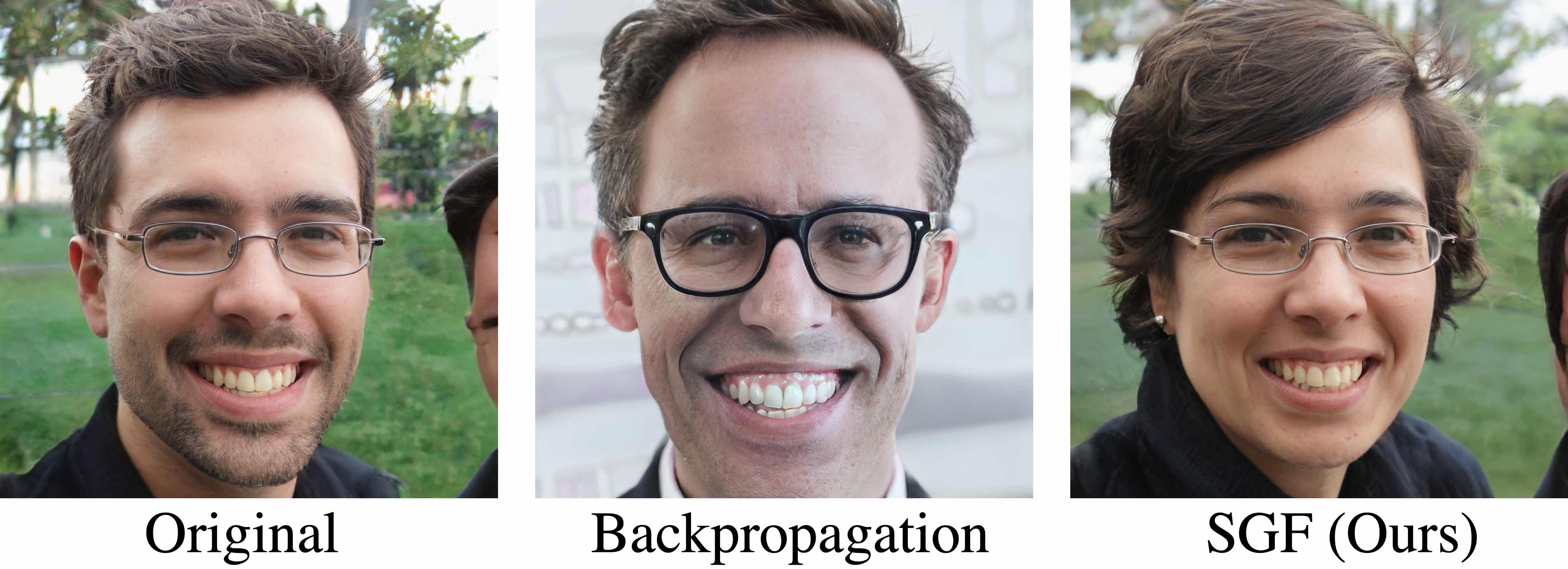}
\end{center}
  \vspace{-10pt}
  \caption{\textbf{Latent code optimization fails to change the gender of the input.} 
  \textbf{Left:} The original face as input. 
  \textbf{Middle:} The result of latent code optimization. 
  We notice that the classifier predicts this face as female. 
  \textbf{Right:} The result of our method.}
\label{fig:compare-optimization}
\end{figure}

As shown in Figure \ref{fig:compare-optimization}, latent code optimization unfortunately does not modify the image as expected. 
We hypothesize that the high degree of non-convexity of the composite function $C \circ G$ leads to this weird behavior.
As a result, gradient-based optimization easily gets stuck in local optima.
A good example of such a local optimum is the face image in the middle of Figure \ref{fig:compare-optimization} which $C$ classifies as a female image.

Another limitation of latent code optimization is that we need to back-propagate $C \circ G$, which may not be possible.
For example, in our Flower-Caption experiments, $C$ is an image captioner followed by a sentence embedding network.
The step of beam search in the caption generation makes it difficult to back-propagate through $C$.

Our method does not suffer from the above two limitations.
We construct a surrogate gradient field of $C \circ G$ to avoid local minima. 
Also, we evaluate $C \circ G$ only in the forward direction, thus avoiding the need to back-propagate $C \circ G$.

\section{Implementation Details of $F$}

\begin{figure}[!t]
\begin{center}
  \includegraphics[width=0.5\linewidth]{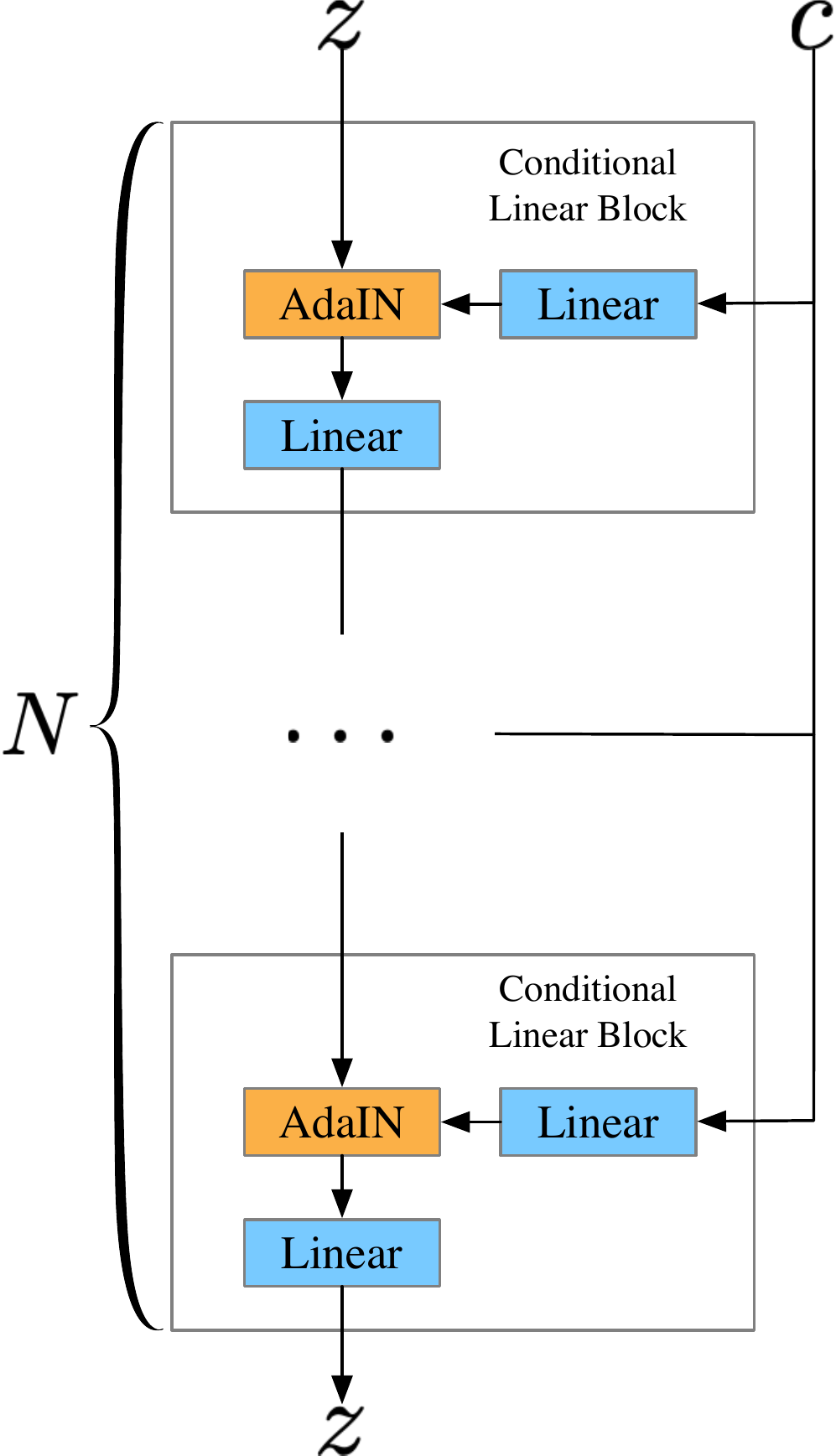}
\end{center}
  \caption{\textbf{The network architecture of the auxiliary mapping network $F$.}
  We use $N=6$ blocks for experiments on Z-Space, $N=15$ blocks for experiments on W-Space. 
  Each conditional linear block contains a fully-connected layer, followed by a AdaIN and a LeakyReLU activation. 
  $c$ is used to calculate the parameters of AdaIN.
  }
\label{fig:supp-network-arch}
\end{figure}

The auxiliary mapping $F$ 
consists of $N$ layers of conditional linear block, as shown in Figure~\ref{fig:supp-network-arch}.
AdaIN represents adaptive normalization introduced in~\cite{huang2017arbitrary}. 
We add a LeakyReLU operation after each AdaIN operation.
The dimension of both hidden features and output features are $512$. 
The length of each latent vector $z$ is $512$ in all experiments, while the length of condition $c$ depends on the experimental settings. The length of condition $c$ is $48$ in FFHQ-Attributes and CelebAHQ-Attributes experiments, $120$ in Anime-KeypointsAttr ($70$ for facial landmarks and $50$ for facial attributes), and $768$ in Flowers-Caption.

\section{Evaluation Details}

Table~\ref{tab:eval-attrs} shows the full attributes list for our facial attributes predictor.
The attributes from No.0 to No.18 are trained using Azure Face API predicted images.
The attributes from No.19 to No.47 are trained using CelebA~\cite{liu2015deep} dataset.
We use all attributes in the experiments on FFHQ and CelebA datasets.

\begin{table}[t]
\caption{\textbf{Attributes list for our facial attribute predictor.}}
\vspace{10pt}
\begin{center}
\resizebox{0.45\columnwidth}{!}{
\begin{tabular}{c|l}
\hline
\bf{No.} & \bf{Attribute}   \\ \hline
0   & Age                   \\ \hline
1   & Gender                \\ \hline
2   & Smile                 \\ \hline
3   & Glasses               \\ \hline
4   & Bald                  \\ \hline
5   & Head\_Roll            \\ \hline
6   & Head\_Yaw             \\ \hline
7   & Head\_Pitch           \\ \hline
8   & Beard                 \\ \hline
9   & Moustache             \\ \hline
10  & Sideburns             \\ \hline
11  & Happiness             \\ \hline
12  & Neutral               \\ \hline
13  & Brown\_Hair           \\ \hline
14  & Black\_Hair           \\ \hline
15  & Blond\_Hair           \\ \hline
16  & Red\_Hair             \\ \hline
17  & Gray\_Hair            \\ \hline
18  & Other\_Hair\_Colors   \\ \hline
19  & 5\_o\_Clock\_Shadow   \\ \hline
20  & Arched\_Eyebrows      \\ \hline
21  & Attractive            \\ \hline
22  & Bags\_Under\_Eyes     \\ \hline
23  & Bangs                 \\ \hline
24  & Big\_Lips             \\ \hline
25  & Big\_Nose             \\ \hline
26  & Blurry                \\ \hline
27  & Bushy\_Eyebrows       \\ \hline
28  & Chubby                \\ \hline
29  & Double\_Chin          \\ \hline
30  & Goatee                \\ \hline
31  & Heavy\_Makeup         \\ \hline
32  & High\_Cheekbones      \\ \hline
33  & Mouth\_Slightly\_Open \\ \hline
34  & Narrow\_Eyes          \\ \hline
35  & No\_Beard             \\ \hline
36  & Oval\_Face            \\ \hline
37  & Pale\_Skin            \\ \hline
38  & Pointy\_Nose          \\ \hline
39  & Receding\_Hairline    \\ \hline
40  & Rosy\_Cheeks          \\ \hline
41  & Straight\_Hair        \\ \hline
42  & Wavy\_Hair            \\ \hline
43  & Wearing\_Earrings     \\ \hline
44  & Wearing\_Hat          \\ \hline
45  & Wearing\_Lipstick     \\ \hline
46  & Wearing\_Necklace     \\ \hline
47  & Wearing\_Necktie      \\ \hline
\end{tabular}
}
\end{center}
\label{tab:eval-attrs}
\end{table}

As we mentioned in the main paper, every manipulation methods have its specific way to control the manipulation intensity, \eg adjusting the length of the vector to apply to control the intensity in InterfaceGAN. 
Users are required to fine-tune the strength of movement along the manipulation path. 
We find that some methods tend to over-modify the image when increasing strength related hyper-parameters, which results in more entangled outputs.
These make it difficult to align the magnitude of editing for each algorithm to make them fairly comparable.
Thus we design the Manipulation Disentanglement Score as a strength-agnostic metric for comparing the disentanglement of image manipulation algorithms.
Table~\ref{tab:ffhq-score} shows detailed evaluation results of Manipulation Disentanglement Score on ``gender'' attribute under the FFHQ-Attributes settings.

\begin{table}[t]
\caption{
    \textbf{Calculation of the MDS of ``gender'' attribute on FFHQ-Attribute dataset.}
    For each method, we increase the manipulation strength (total iteration $n$ for SGF, vector length $\mu$ for InterfaceGAN) until the manipulation accuracy reaches $1$ or accumulated MDS starts to decrease, and choose the maximum accumulated MDS as the final MDS.
}
\vspace{-5pt}
\begin{center}
\resizebox{1\columnwidth}{!}{
\begin{tabular}{l cccc}
\hline
 & \multicolumn{2}{c}{\bf{Manipulation}}
 & \multicolumn{1}{c}{\bf{Accumulated}} 
 & \multicolumn{1}{c}{\bf Harmonic Means of}\\
\bf{Method} &  \bf{Acc.} & \bf{Disent.} & \bf{MDS} & \bf{Acc. \& Disent.}  \\
\hline  
SGF $, n=5$ & 0.18 & 0.986 & 0.179 & 0.304 \\
SGF $, n=10$ & 0.48 & 0.915 & 0.464 & 0.630 \\
SGF $, n=15$ & 0.79 & 0.890  & 0.744 & 0.837 \\
SGF $, n=20$ & 0.93 & 0.872 & 0.867 & 0.900 \\
SGF $, n=25$ & 0.99 & 0.859 & \bf0.919 & \bf0.920 \\
SGF $, n=30$ & 0.98 & 0.842 & 0.910 & 0.906 \\
\hline
InterfaceGAN $, \mu=0.25$& 0.13 & 0.993 & 0.129 & 0.230 \\
InterfaceGAN $, \mu=0.5$& 0.32 & 0.942 & 0.312 & 0.478 \\
InterfaceGAN $, \mu=0.75$& 0.41 & 0.883 & 0.394 & 0.560 \\
InterfaceGAN $, \mu=1.0$& 0.55 & 0.822 & 0.513 & 0.659 \\
InterfaceGAN $, \mu=2.0$& 0.85 & 0.612 & 0.728 & \bf0.712 \\
InterfaceGAN $, \mu=3.0$& 0.99 & 0.469 & 0.804 & 0.636 \\
InterfaceGAN $, \mu=4.0$& 1.00    & 0.398 & \bf0.808 & 0.569 \\
\hline
\end{tabular}
}
\end{center}
\label{tab:ffhq-score}
\end{table}

\section{Additional Results on FFHQ-Attributes}

\begin{figure}[!t]
\begin{center}
  \includegraphics[width=1\linewidth]{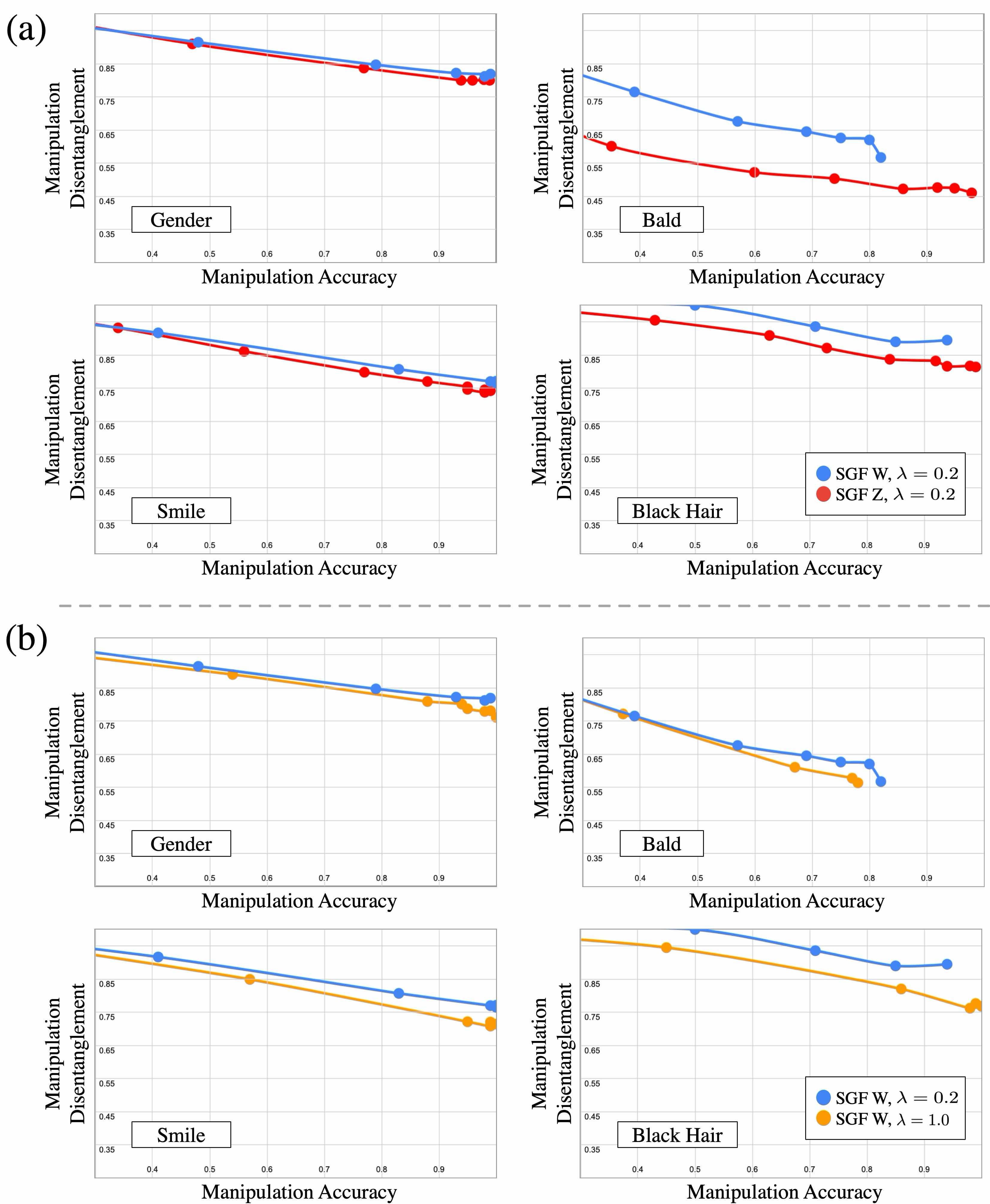}
\end{center}
  \caption{\textbf{The MDS of ablation study on our method in the FFHQ-Attributes dataset.}
  (a) The MDC of our ablation study in different latent spaces.
  (b) The MDC of our ablation study using different step size $\lambda$.
}
\label{fig:supp-ablation}
\end{figure}

\begin{figure*}[t]
\begin{center}
  \centering
  \includegraphics[width=0.9\linewidth]{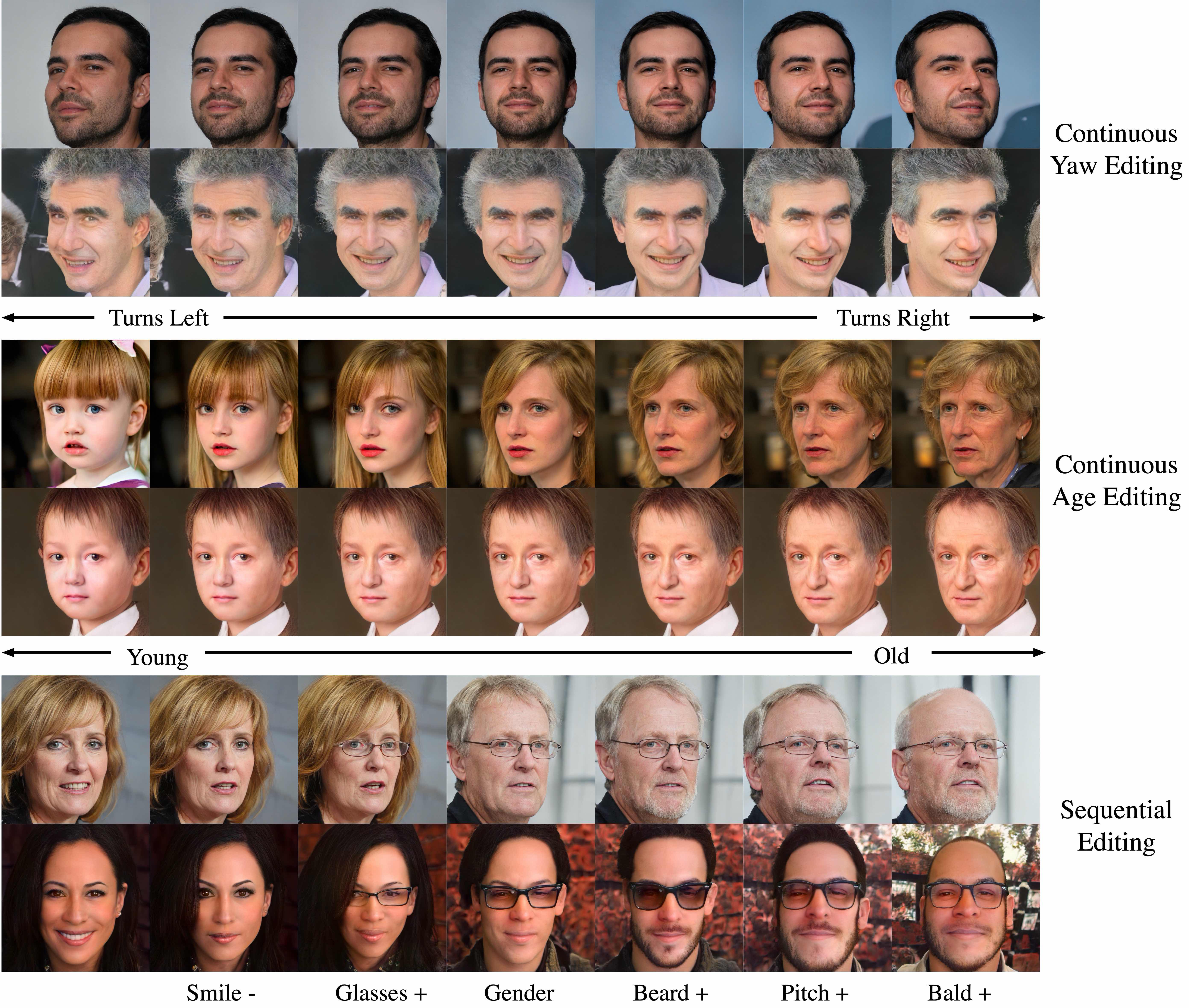}
\end{center}
 \vspace{-1.5em}
  \caption{\textbf{Additional results on the FFHQ-Attributes dataset.} We edit face images generated by GANs in odd rows and edit real-life images projected to the latent space of GANs in even rows.
  }
\label{fig:ffhq-results}
\vspace{-1em}
\end{figure*}

\begin{figure*}[t!]
\begin{center}
  \centering
  \includegraphics[width=0.85\linewidth]{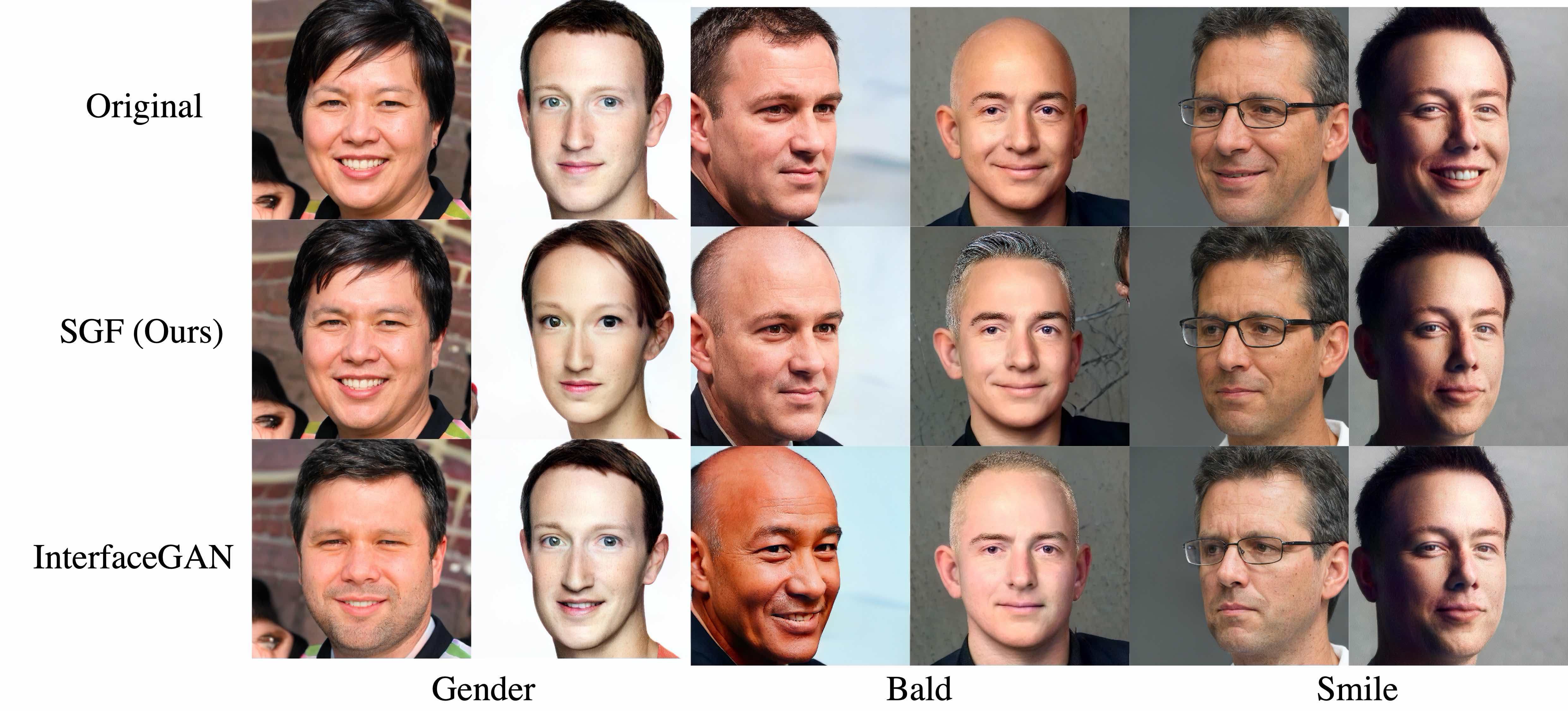}
\end{center}
\vspace{-1.5em}
  \caption{\textbf{Part of the samples used in the user study.} The original images in odd columns are generated by GANs while those in even columns are real-life images projected to the latent space of GANs.
  }
\label{fig:user-study}
 \vspace{-1em}
\end{figure*}

Figure~\ref{fig:ffhq-results} shows the additional results of attributes manipulation in FFHQ-Attributes dataset.
The first two rows show results of editing facial orientation by using different values of ``yaw''. 
The second two rows show continuous editing results of ``age''. 
Finally, the last two rows are results of sequential editing using SGF.
In Figure~\ref{fig:user-study}, we also show some samples used in our user study.

\section{Additional Results on CelebAHQ-Attributes}

\begin{figure*}[t]
\begin{center}
  \centering
  \vspace{50pt}
  \includegraphics[width=1\textwidth]{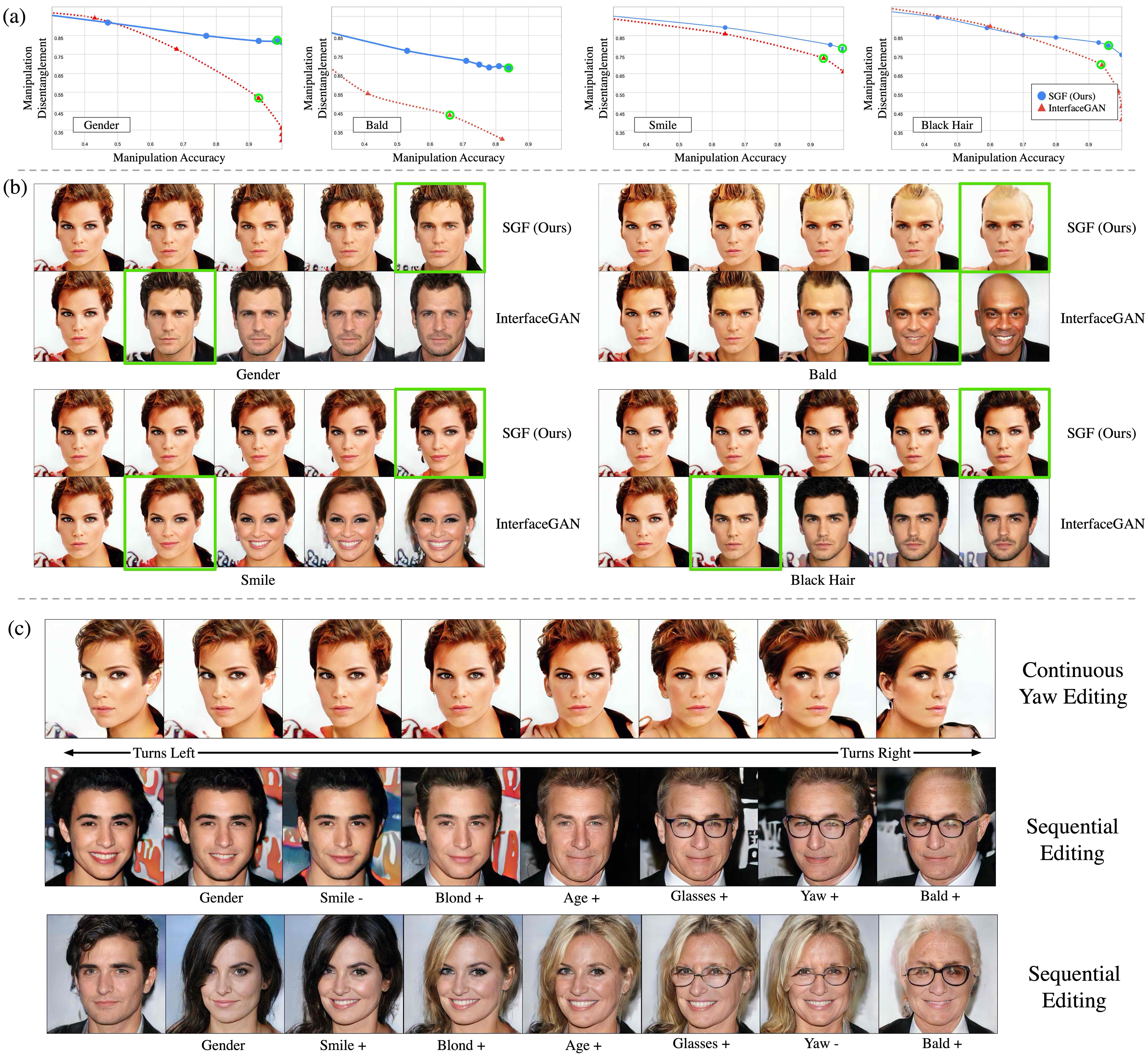}
  \caption{\textbf{Facial attribute editing results in the CelebAHQ-Attributes dataset.} 
  (a) The MDC of InterfaceGAN and our method on several attributes.
  Green circles highlight the image that has the highest harmonic mean of accuracy and disentanglement along the curve.
  (b) Manipulation process of InterfaceGAN and our method.
  Green boxes highlight the images with the highest harmonic mean of accuracy and disentanglement during the manipulation.
  (c) More results of sequential editing using our method.
  \vspace{50pt}
  }\label{fig:supp-celeba}
\end{center}
\end{figure*}

The MDCs of CelebAHQ-Attributes dataset are shown in Figure~\ref{fig:supp-celeba}(a).
Green circles highlight the image that has the highest harmonic mean of accuracy and disentanglement along the curve.

We show more comparisons in Figure~\ref{fig:supp-celeba}(b) to illustrate the effect of different hyper-parameters on the results of each method.
For SGF, the hyper-parameter refers to the max step number $n$, while for InterfaceGAN it is the magnitude of displacement in the direction of condition.
Both can be interpreted as the process of increasing the intensity of manipulation.
Green boxes highlight the results that use the corresponding highlighted hyper-parameters in Figure~\ref{fig:supp-celeba}(a).
As we can see, similar to the results in FFHQ-Attributes, our method shows higher disentanglement on each attribute, changing the target attribute while keeping other attributes intact during the manipulation process.

Figure~\ref{fig:supp-celeba}(c) shows the results of continuous attribute adjustment.
The first row shows results of gradual adjustment of the yaw attribute, and the following rows show sequential adjustments of several attributes.
For each row, we observe smooth translation from the original image to the target image, which facilitates an overall more realistic editing sequence.

\section{Additional Results on Anime-KeypointsAttr}

\begin{figure}[!t]
\begin{center}
  \includegraphics[width=1\linewidth]{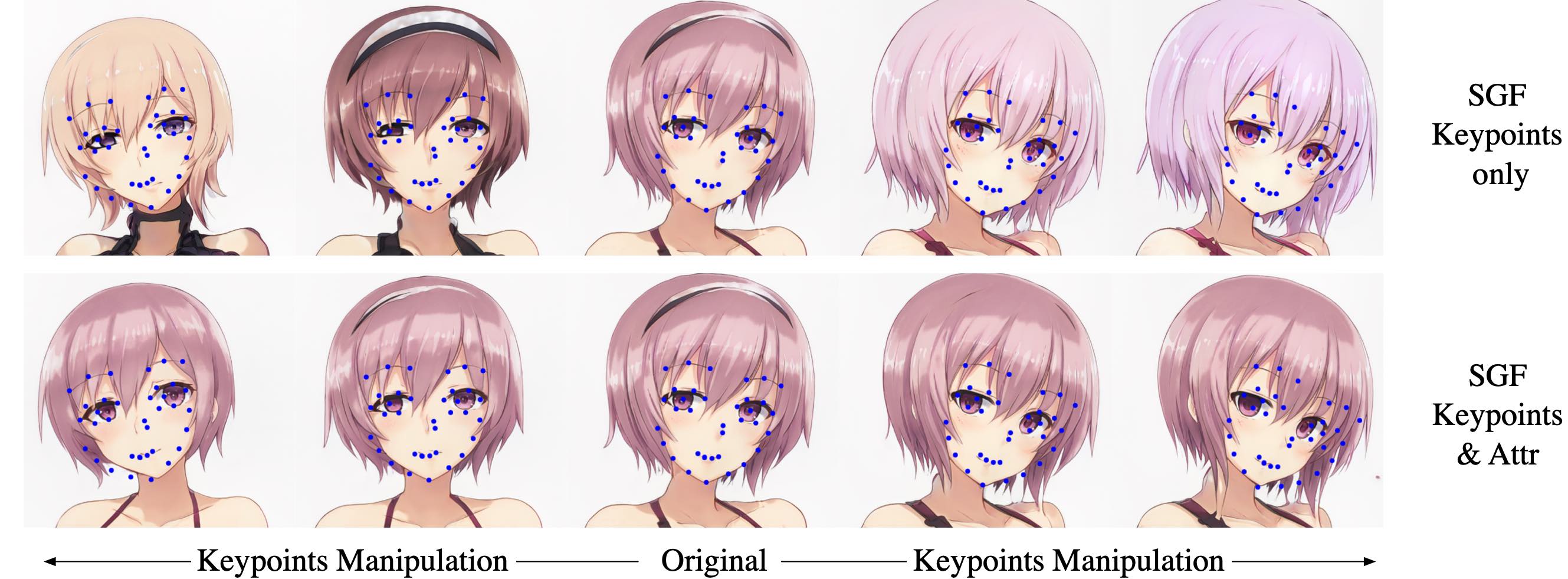}
\end{center}
  \caption{\textbf{Comparison on keypoint manipulation results on Anime-Keypoints and Anime-KeypointsAttr.}
The first row shows the manipulation results of SGF conditioned on keypoints only. 
The second row shows the results conditioned on both keypoints and attributes.
Additional controllablilty on attributes (e.g. hair color) ensures that they are consistent when keypoints change.
}
\label{fig:keypoints-attr-comp}
\end{figure}

We found that only using keypoints as the control condition could cause undesired changes in results, as shown in the first row in Figure~\ref{fig:keypoints-attr-comp}.
Thus the keypoints prediction is concatenated with the first $50$ attributes from {illustration2vec} \cite{saito2015illustration2vec} classifier as the final \textbf{KeypointsAttr} condition for keypoints manipulation experiments.
Since the predicted attributes in KeypointsAttr contain hair color information, adding these attributes to training conditions can alleviate the undesired changes in results.
For the $F$ trained with both keypoints and attributes (the second row in Figure~\ref{fig:keypoints-attr-comp}), our method successfully maintains the hair color unchanged after the manipulation, indicating that adding conditioning variables can encourage our method to perform better disentanglement.

\begin{figure*}[t!]
\begin{center}
  \centering
  \includegraphics[width=1\textwidth]{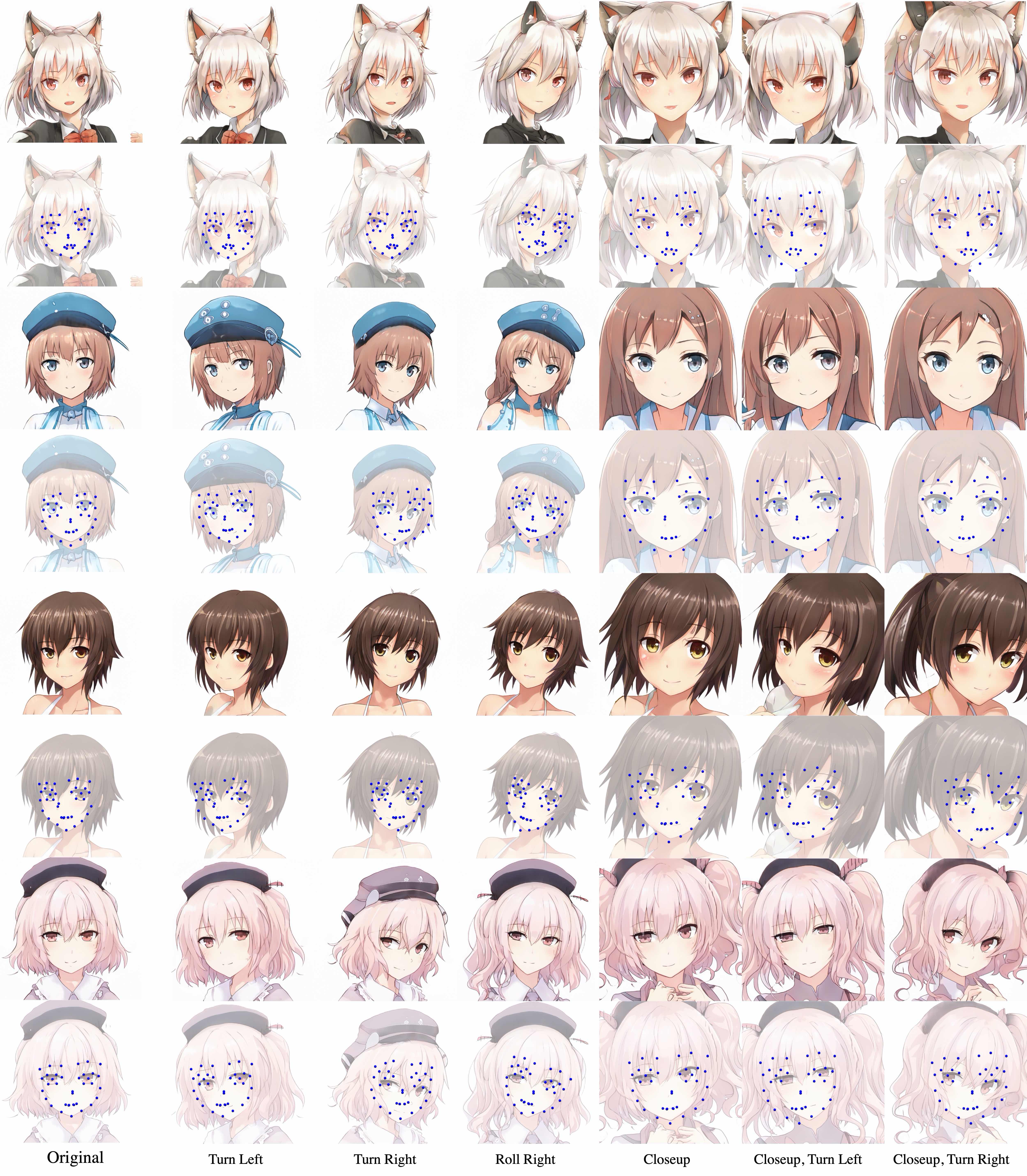}
  \caption{\textbf{Keypoints editing on the Anime-KeypointsAttr dataset.} 
  Odd rows show the results of our keypoints manipulation and even rows are the corresponding target keypoint conditions.
}\label{fig:supp-anime}
\end{center}
\end{figure*}

Figure~\ref{fig:supp-anime} shows additional manipulation results in Anime-KeypointsAttr dataset.
We use our method to edit the head poses and zoom levels of generated anime faces (columns two through five). 
In addition, we show some sequential editing of zoom level and head poses (last two columns).

\section{Additional Results on Flower-Caption}

\begin{figure*}[t!]
\begin{center}
  \centering
  \vspace{20pt}
  \includegraphics[width=1\textwidth]{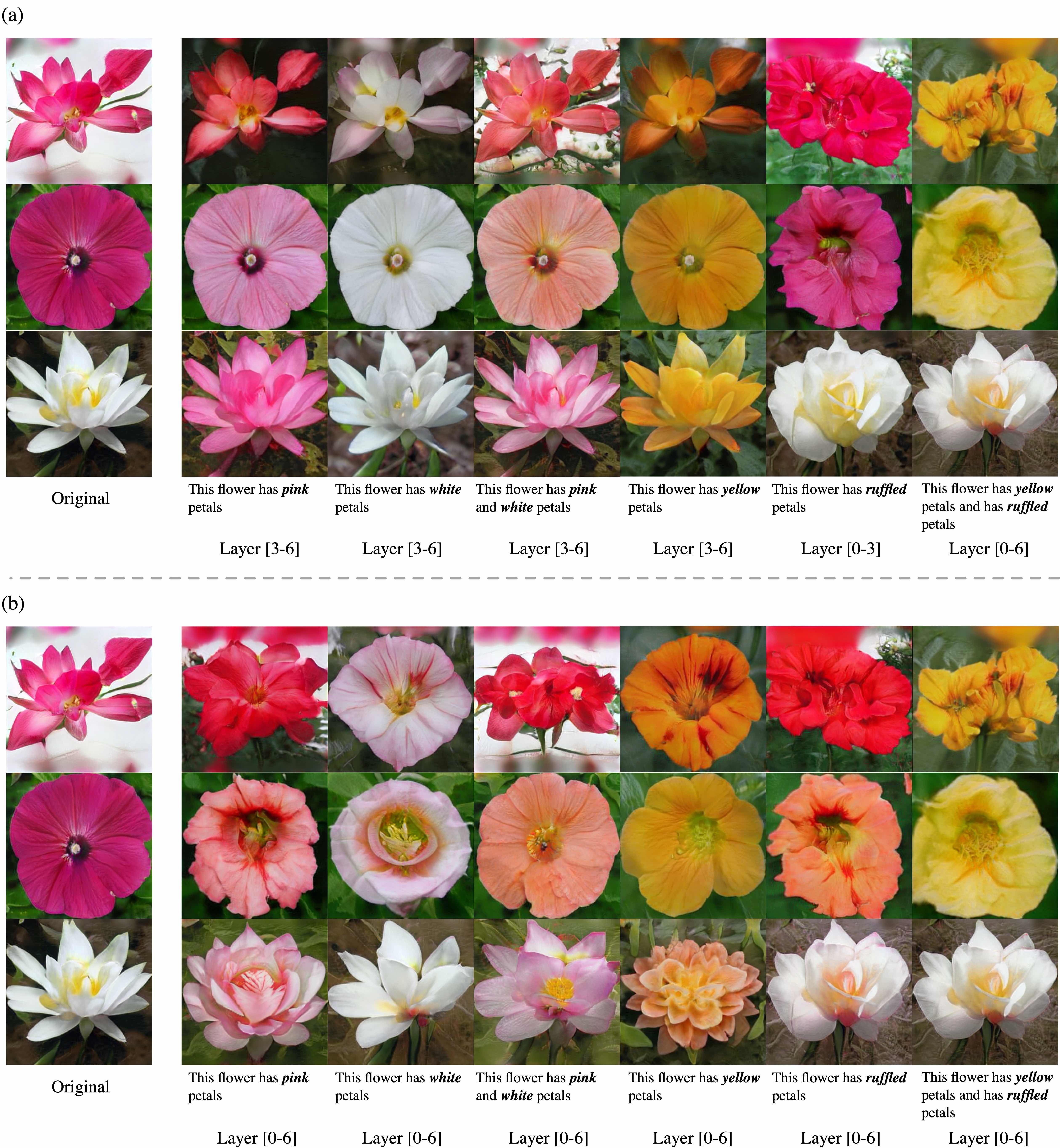}
  \caption{\textbf{Manipulation by captions in Flowers-Caption dataset.}
  (a) Manipulation results on Flower-Caption dataset. 
   We limit the latent space manipulation to certain layers when editing either color or shapes, and apply to all layers when conditioning on both color and shape.
  The layers used for each manipulation are noted at the bottom of each column.
  (b) Manipulation results without layer-wise manipulation in (a).
}\label{fig:supp-flowers}
\end{center}
\end{figure*}

Figure~\ref{fig:supp-flowers}(a) shows additional results of Flower-Caption experiments using the same configuration as that used in the paper.
Note that unlike Anime-KeypointsAttr experiments, we do not have additional conditioning variables to keep the shape of the flower unchanged when editing the color. 
We limit latent space manipulation to apply on the top four layers for color manipulations, the bottom four layers for shape manipulations, while imposing no limitation for editing that involves changes of both color and shape.
Figure~\ref{fig:supp-flowers}(b) shows the results using the same inputs as Figure~\ref{fig:supp-flowers}(a) without limiting the latent space manipulations to apply on specific layers in StyleGAN2.
We observe that conditioning on colors using all layers might result in unwanted shape changes, and vise versa.

\section{Ablation Study}\label{sec:ablation}

We conduct an ablation study on latent space of GANs and the step size in our algorithm.
Figure~\ref{fig:supp-ablation}(a) shows the MDC of our method in either Z-space or W-space of the StyleGAN model trained on the FFHQ-Attributes dataset.
Given the same attribute to manipulate, our model performs better in the W-space of StyleGAN2~\cite{karras2020analyzing} than in the Z-space.
This shows that our model can benefit from a more disentangled latent space~\cite{karras2019style}.
Figure~\ref{fig:supp-ablation}(b) shows the MDC of our method using different step size $\lambda$.
Using a smaller step size improves accuracy, resulting in a better MDC shape. 
However, using step sizes that are too small leads to slow convergences.
Here we omit the MDC of $\lambda = 0.02$ and below because our method under such settings reachs the maximum iteration with an accuracy lower than $0.3$ most of the time.

In table \ref{tab:ablation}, we report quantitative results of SGF under different hyper-parameter settings on FFHQ-Attributes dataset.
We set the max iteration number $n=50$ and examine performances under different step sizes $\lambda$.
We find that decreasing the step size from $1$ to $0.2$ could result in a better performance.
However, a further decrease to $0.02$ significantly slows down the convergences, thus shows inferior results in MDS.
We also test performances on Z-space, we observe performance drop for all step sizes comparing with W-space. 
This suggests that our method can benefit from using a better disentangled latent space.

\begin{table}[t]
\caption{\textbf{MDS comparison on FFHQ for ablation study,}
    evaluated using different step size $\lambda$ for SGF on Z-space and W-space of StyleGAN2. 
    \vspace{-5pt}
} 
\begin{center}
\resizebox{1\columnwidth}{!}{
\begin{tabular}{lcccccc}
\hline
\bf{Method} & \bf{Gender}  &  \bf{Bald} & \bf{Smile} & \bf{Black Hair} &  \bf{Overall}\\
\hline 
\ourmethod Z, $\lambda = 1$ & 0.857 & 0.519 &  0.846 &  0.886 & 0.777 \\ 
\ourmethod Z, $\lambda = 0.2$ & 0.901 & 0.579 & 0.839 & 0.909 & 0.807 \\ 
\ourmethod Z, $\lambda = 0.02$ & 0.353 & 0.039 & 0.258 &  0.140 & 0.198 \\ 
\hline
\ourmethod W, $\lambda = 1$ & 0.889 & 0.587 & 0.859 &  0.911 & 0.812 \\ 
\ourmethod W, $\lambda = 0.2$ & \bf0.919 & \bf0.590 &\bf0.884 & \bf0.955 & \bf0.837 \\ 
\ourmethod W, $\lambda = 0.02$ & 0.209 & 0.125 & 0.649 &  0.240 & 0.306 \\ 
\hline
\end{tabular}
}
\end{center}
\vspace{-5pt}
\label{tab:ablation}
\end{table}

\section{Non-Linear Path of SGF} \label{sec:nonlinear}

\begin{figure*}[t]
\begin{center}
  \centering
  \includegraphics[width=1\linewidth]{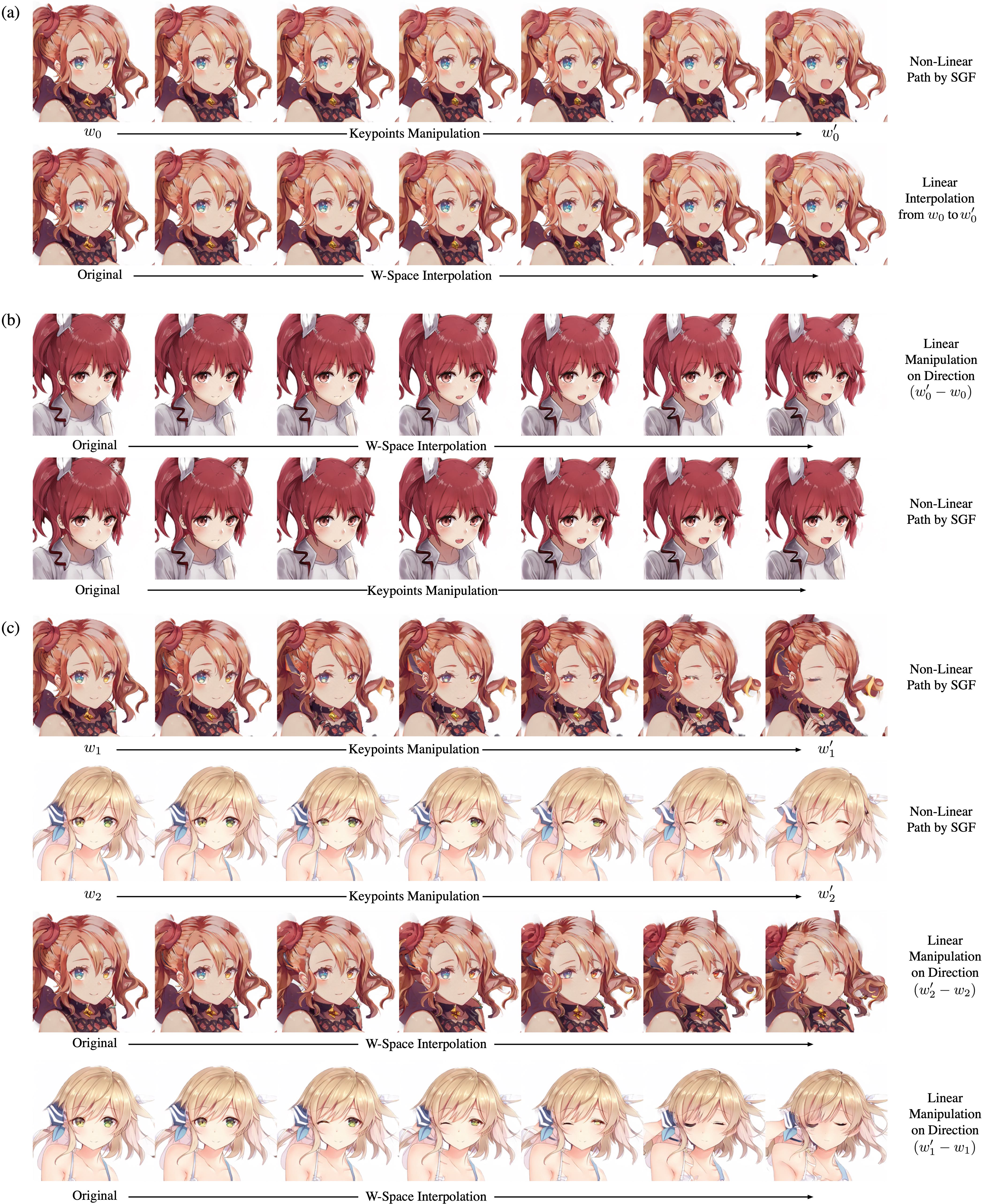}
\end{center}
  \caption{\textbf{
  Comparing the non-linear path of SGF and its linear interpolations on
  Anime-KeypointsAttr dataset.} 
  (a) Comparing the non-linear path of SGF and its linear interpolations.
  (b) Using the manipulation direction in (a) to control another latent sample.
  (c) Use SGF to get the ``eye closed'' directions, and then exchange the manipulation direction of two latent samples.
  Refer to Section~\ref{sec:nonlinear} for the details.}
\label{fig:linearity-comp}
\vspace{-1em}
\end{figure*}

Compared with previous approaches, our SGF model has non-linear, position-variant properties when manipulating latent codes.
Figure~\ref{fig:linearity-comp}(a) shows the editing path of SGF and its linear interpolation on mouth keypoints editing.
Specifically, the non-linear results are obtained by setting different iteration limit $n$ on SGF, while the linear interpolation is interpolation on latent space from the starting $w_0$ to final $w_0'$ calculated by SGF.
We notice that the linear interpolation makes a close approximation to the original non-linear path.

Applying the ``mouth open'' direction $(w_0' - w_0)$ to another latent point also achieves similar results to the non-linear path from SGF, as shown in Figure~\ref{fig:linearity-comp}(b).
However, such transferability does not apply to every manipulation.
As shown in Figure~\ref{fig:linearity-comp}(c), linearly applying the ``eyes closed'' direction obtained from SGF to other latent points generates results inferior to the original non-linear results. Although both linear manipulation operations close the eyes of anime charactors, they also introduce unwanted mouth manipulation (row 3) and unnatural editing (row 4) in the final results.

Overall, we can simplify the control direction obtained by SGF to linear control without making significant sacrifices, and in some cases, such linear direction can be applied to other samples.  To ensure making precise and disentangled modification on given results, however, needs to rely on the non-linear path of SGF. 

\section{Running Time of SGF}

\begin{table}[!t]
\caption{\textbf{Running time of SGF on different datasets (in sec.)}.
    The running time of SGF depends on the re-estimation step $c_i = C(G(z_i))$, using a larger generator and condition predictor would result in a longer running time.}
\vspace{-5pt}
\begin{center}
\resizebox{0.7\columnwidth}{!}{
\begin{tabular}{lccc}
\hline
\bf{Method} & \bf{SGF}  &  \bf{SGF (fast ver.)}\\
\hline
FFHQ-Attributes & $2.89$ & $0.304$ \\
Anime-KeypointsAttr & $6.54$ &  $0.332$ \\  
Flower-Captions & $11.73$ &  $0.366$ \\  
\hline
\end{tabular}
}
\end{center}
\label{tab:running-time}
\end{table}

Table~\ref{tab:running-time} shows the running time (in second) of SGF on different settings, averaged from $100$ samples.
The running time of SGF (\ie, running time of Algorithm~1) largely depends on the running time of inferencing condition of new latent code in each step $c^{(i)} = C(G(z^{(i)}))$.
We observe faster running time in FFHQ-Attributes experiments, which uses a facial attributes classifier  that has a much simpler structure compared to the keypoint attribute predictor in Anime-KeypointsAttr.
The $C$ in Flower-Captions consists of an image captioner and a sentence embedding encoder, making the overall running time much longer than other settings.

A trick to save time when running Algorithm~1 is to replace the inference step $c^{(i)} = C(G(z^{(i)}))$ by $c^{(i)}= i\delta_c$ after the first inference.
As a result, the Algorithm~1 runs at approximately a constant $0.4$ second, shown as SGF (fast ver.) in Table~\ref{tab:running-time}.
This trick would make some scarification on manipulation performance due to the estimation error of Auxiliary Mapping $F$.

\clearpage
\clearpage

{\small
\bibliographystyle{ieee_fullname}

\begin{thebibliography}{10}\itemsep=-1pt

\bibitem{abdal2019image2stylegan}
Rameen Abdal, Yipeng Qin, and Peter Wonka.
\newblock Image2stylegan: How to embed images into the stylegan latent space?
\newblock In {\em Proceedings of ICCV}, 2019.

\bibitem{abdal2020styleflow}
Rameen Abdal, Peihao Zhu, Niloy Mitra, and Peter Wonka.
\newblock Styleflow: Attribute-conditioned exploration of stylegan-generated
  images using conditional continuous normalizing flows.
\newblock {\em arXiv preprint arXiv:2008.02401}, 2020.

\bibitem{arjovsky2017wasserstein}
Martin Arjovsky, Soumith Chintala, and L{\'e}on Bottou.
\newblock Wasserstein generative adversarial networks.
\newblock In {\em Proceedings of ICML}, 2017.

\bibitem{bau2019semantic}
David Bau, Hendrik Strobelt, William Peebles, Jonas Wulff, Bolei Zhou, Jun-Yan
  Zhu, and Antonio Torralba.
\newblock Semantic photo manipulation with a generative image prior.
\newblock {\em ACM Transactions on Graphics (TOG)}, 38(4):59, 2019.

\bibitem{behrmann2019invertible}
Jens Behrmann, Will Grathwohl, Ricky~TQ Chen, David Duvenaud, and
  J{\"o}rn-Henrik Jacobsen.
\newblock Invertible residual networks.
\newblock In {\em Proceedings of ICML}, 2019.

\bibitem{brock2018large}
Andrew Brock, Jeff Donahue, and Karen Simonyan.
\newblock Large scale gan training for high fidelity natural image synthesis.
\newblock In {\em Proceedings of ICLR}, 2018.

\bibitem{cao2018vggface2}
Qiong Cao, Li Shen, Weidi Xie, Omkar~M Parkhi, and Andrew Zisserman.
\newblock Vggface2: A dataset for recognising faces across pose and age.
\newblock In {\em Proceedings of IEEE International Conference on Automatic
  Face \& Gesture Recognition}, 2018.

\bibitem{goetschalckx2019ganalyze}
Lore Goetschalckx, Alex Andonian, Aude Oliva, and Phillip Isola.
\newblock Ganalyze: Toward visual definitions of cognitive image properties.
\newblock In {\em Proceedings of CVPR}, 2019.

\bibitem{goodfellow2014generative}
Ian Goodfellow, Jean Pouget-Abadie, Mehdi Mirza, Bing Xu, David Warde-Farley,
  Sherjil Ozair, Aaron Courville, and Yoshua Bengio.
\newblock Generative adversarial nets.
\newblock In {\em Proceedings of NIPS}, 2014.

\bibitem{gulrajani2017improved}
Ishaan Gulrajani, Faruk Ahmed, Martin Arjovsky, Vincent Dumoulin, and Aaron~C
  Courville.
\newblock Improved training of wasserstein gans.
\newblock In {\em Proceedings of NIPS}, 2017.

\bibitem{harkonen2020ganspace}
Erik H{\"a}rk{\"o}nen, Aaron Hertzmann, Jaakko Lehtinen, and Sylvain Paris.
\newblock Ganspace: Discovering interpretable gan controls.
\newblock {\em arXiv preprint arXiv:2004.02546}, 2020.

\bibitem{horn_johnson_2012}
Roger~A. Horn and Charles~R. Johnson.
\newblock {\em Matrix Analysis}.
\newblock Cambridge University Press, 2 edition, 2012.

\bibitem{hu2018squeeze}
Jie Hu, Li Shen, and Gang Sun.
\newblock Squeeze-and-excitation networks.
\newblock In {\em Proceedings of CVPR}, 2018.

\bibitem{huang2017arbitrary}
Xun Huang and Serge Belongie.
\newblock Arbitrary style transfer in real-time with adaptive instance
  normalization.
\newblock In {\em Proceedings of ICCV}, 2017.

\bibitem{isola2016image}
Phillip Isola, Jun-Yan Zhu, Tinghui Zhou, and Alexei~A Efros.
\newblock Image-to-image translation with conditional adversarial networks.
\newblock In {\em Proceedings of CVPR}, 2017.

\bibitem{jahanian2019steerability}
Ali Jahanian, Lucy Chai, and Phillip Isola.
\newblock On the" steerability" of generative adversarial networks.
\newblock In {\em Proceedings of ICLR}, 2019.

\bibitem{jin2017towards}
Yanghua Jin, Jiakai Zhang, Minjun Li, Yingtao Tian, Huachun Zhu, and Zhihao
  Fang.
\newblock Towards the automatic anime characters creation with generative
  adversarial networks.
\newblock {\em arXiv preprint arXiv:1708.05509}, 2017.

\bibitem{karras2018progressive}
Tero Karras, Timo Aila, Samuli Laine, and Jaakko Lehtinen.
\newblock Progressive growing of gans for improved quality, stability, and
  variation.
\newblock In {\em Proceedings of ICLR}, 2018.

\bibitem{karras2019style}
Tero Karras, Samuli Laine, and Timo Aila.
\newblock A style-based generator architecture for generative adversarial
  networks.
\newblock In {\em Proceedings of CVPR}, 2019.

\bibitem{karras2020analyzing}
Tero Karras, Samuli Laine, Miika Aittala, Janne Hellsten, Jaakko Lehtinen, and
  Timo Aila.
\newblock Analyzing and improving the image quality of stylegan.
\newblock In {\em Proceedings of CVPR}, 2020.

\bibitem{kingma2015adam}
Diederik~P Kingma and Jimmy Ba.
\newblock Adam: A method for stochastic optimization.
\newblock In {\em Proceedings of ICLR}, 2015.

\bibitem{ledig2017photo}
Christian Ledig, Lucas Theis, Ferenc Husz{\'a}r, Jose Caballero, Andrew
  Cunningham, Alejandro Acosta, Andrew Aitken, Alykhan Tejani, Johannes Totz,
  Zehan Wang, et~al.
\newblock Photo-realistic single image super-resolution using a generative
  adversarial network.
\newblock In {\em Proceedings of CVPR}, 2017.

\bibitem{liu2015deep}
Ziwei Liu, Ping Luo, Xiaogang Wang, and Xiaoou Tang.
\newblock Deep learning face attributes in the wild.
\newblock In {\em Proceedings of ICCV}, 2015.

\bibitem{miyato2019spectral}
Takeru Miyato, Toshiki Kataoka, Masanori Koyama, and Yuichi Yoshida.
\newblock Spectral normalization for generative adversarial networks.
\newblock In {\em Proceedings of ICLR}, 2018.

\bibitem{nilsback2008automated}
Maria-Elena Nilsback and Andrew Zisserman.
\newblock Automated flower classification over a large number of classes.
\newblock In {\em 2008 Sixth Indian Conference on Computer Vision, Graphics \&
  Image Processing}, pages 722--729. IEEE, 2008.

\bibitem{plumerault2019controlling}
Antoine Plumerault, Herv{\'e} Le~Borgne, and C{\'e}line Hudelot.
\newblock Controlling generative models with continuous factors of variations.
\newblock In {\em Proceedings of ICLR}, 2019.

\bibitem{radford2015unsupervised}
Alec Radford, Luke Metz, and Soumith Chintala.
\newblock Unsupervised representation learning with deep convolutional
  generative adversarial networks.
\newblock {\em arXiv preprint arXiv:1511.06434}, 2015.

\bibitem{reed2016generative}
Scott Reed, Zeynep Akata, Xinchen Yan, Lajanugen Logeswaran, Bernt Schiele, and
  Honglak Lee.
\newblock Generative adversarial text to image synthesis.
\newblock In {\em Proceedings of ICML}, 2016.

\bibitem{reimers2019sentence}
Nils Reimers and Iryna Gurevych.
\newblock Sentence-bert: Sentence embeddings using siamese bert-networks.
\newblock In {\em Proceedings of EMNLP}, 2019.

\bibitem{saito2015illustration2vec}
Masaki Saito and Yusuke Matsui.
\newblock Illustration2vec: a semantic vector representation of illustrations.
\newblock In {\em SIGGRAPH Asia 2015 Technical Briefs}. 2015.

\bibitem{shen2020interpreting}
Yujun Shen, Jinjin Gu, Xiaoou Tang, and Bolei Zhou.
\newblock Interpreting the latent space of gans for semantic face editing.
\newblock In {\em Proceedings of CVPR}, 2020.

\bibitem{voynov2020unsupervised}
Andrey Voynov and Artem Babenko.
\newblock Unsupervised discovery of interpretable directions in the gan latent
  space.
\newblock In {\em Proceedings of ICML}, 2020.

\bibitem{xu2015show}
Kelvin Xu, Jimmy Ba, Ryan Kiros, Kyunghyun Cho, Aaron Courville, Ruslan
  Salakhudinov, Rich Zemel, and Yoshua Bengio.
\newblock Show, attend and tell: Neural image caption generation with visual
  attention.
\newblock In {\em Proceedings of ICML}, 2015.

\bibitem{zhang2016stackgan}
Han Zhang, Tao Xu, Hongsheng Li, Shaoting Zhang, Xiaolei Huang, Xiaogang Wang,
  and Dimitris Metaxas.
\newblock Stackgan: Text to photo-realistic image synthesis with stacked
  generative adversarial networks.
\newblock In {\em Proceedings of ICCV}, 2016.

\bibitem{zhao2020differentiable}
Shengyu Zhao, Zhijian Liu, Ji Lin, Jun-Yan Zhu, and Song Han.
\newblock Differentiable augmentation for data-efficient gan training.
\newblock In {\em Proceedings of NIPS}, 2020.

\bibitem{zhu2017unpaired}
Jun-Yan Zhu, Taesung Park, Phillip Isola, and Alexei~A Efros.
\newblock Unpaired image-to-image translation using cycle-consistent
  adversarial networks.
\newblock In {\em Proceedings of ICCV}, 2017.

\end{thebibliography}

}

\end{document}